\pdfoutput=1

\documentclass[11pt]{article}

\usepackage[]{acl}
\usepackage{booktabs}

\usepackage{times}
\usepackage{latexsym}
\usepackage{amsmath}
\usepackage{tabularx}
\usepackage{footnote}
\makesavenoteenv{tabular}
\makesavenoteenv{table}

\newcolumntype{B}{X}
\newcolumntype{s}{>{\hsize=.5\hsize}X}

\usepackage[T1]{fontenc}

\usepackage[utf8]{inputenc}

\usepackage{microtype}

\usepackage{inconsolata}
\usepackage[size=tiny]{todonotes}
\usepackage{subcaption}

\newcommand{\se}[1]{\textcolor{black}{#1}}

\newif\ifcomments
\commentstrue
\ifcomments
    \providecommand\rotem[1]{\textcolor{green}{[rotem: #1]}}
    \providecommand\todo[1]{\textcolor{red}{[TODO: #1]}}
\else
    \providecommand{\rotem}[1]{}
    \providecommand{\todo}[1]{}
\fi
\title{The Eval4NLP 2023 Shared Task on \\
Prompting Large Language Models as Explainable Metrics \\~\\\includegraphics[width=0.110\textwidth]{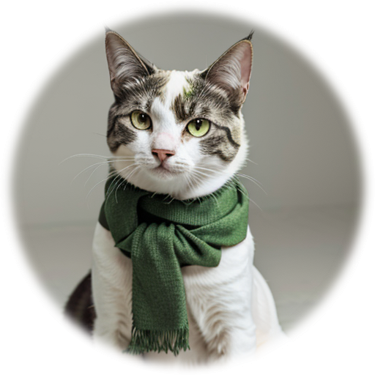}}

\author{\textbf{Christoph Leiter}$^{*}$, 
\textbf{Juri Opitz}$^{\dagger}$,
\textbf{Daniel Deutsch}$^{\ddagger}$,\\ 
\textbf{Yang Gao}$^{\diamond}$
\textbf{Rotem Dror$^{\dagger\dagger}$},
\textbf{Steffen Eger$^{*}$}
\\
$*$ Bielefeld University, Germany \quad
$\dagger$ Heidelberg University, Germany \\
$\ddagger$ Google, US \quad
$\diamond$ Google Research, UK \quad 
$\dagger\dagger$ University of Haifa, Israel
\\
\texttt{christoph.leiter@uni-bielefeld.de} \quad
\texttt{opitz.sci@gmail.com} \\
\texttt{dandeutsch@google.com} \quad
\texttt{gaostayyang@google.com} \quad \\
\texttt{rdror@is.haifa.ac.il} \quad
\texttt{steffen.eger@uni-bielefeld.de}
}

\newcommand{\cl}[1]{\textcolor{black}{#1}}
\newcommand{\cll}[1]{\textcolor{black}{#1}}

\begin{document}
\maketitle
\begin{abstract}
With an increasing number of parameters and pre-training data, 
generative large language models (LLMs) 
have shown remarkable capabilities to solve tasks with minimal or no task-related examples. 
Notably, LLMs have been successfully employed as evaluation metrics in text generation tasks. 
\cll{Within this context, we introduce the Eval4NLP 2023 shared task that asks participants to explore \textit{prompting} and \textit{score extraction} for machine translation (MT) 
and summarization evaluation. Specifically, we propose a novel competition setting in which we select a list of allowed LLMs and disallow fine-tuning to ensure a focus on prompting.
We present an overview of participants' approaches and evaluate them on a new reference-free test set spanning three language pairs for MT and a summarization dataset. Notably, despite the task's restrictions, the best-performing systems %are achieving 
achieve 
results on par with or even surpassing recent reference-free metrics developed using larger models, including GEMBA and Comet-Kiwi-XXL for %machine translation.}
MT.} 
Finally, as a separate track, we perform a small-scale human evaluation of the plausibility of explanations given by the LLMs.\footnote{ We make parts of our code and datasets available: \url{https://github.com/eval4nlp/SharedTask2023/tree/main}}

\end{abstract}

\section{Introduction}
\label{sec:introduction}
The ChatGPT revolution in late 2022 has ignited 
a wide public and scientific debate about the possibilities (and limitations) of generative AI in various fields and application scenarios \citep{leiter2023chatgpt,Eger2023NLLGQA}, including education \citep{Halaweh2023ChatGPTIE}, logic \citep{liu2023evaluating}, medicine \citep{Dave2023ChatGPTIM}, math \citep{frieder2023mathematical}, programming \citep{rozière2023code} and science \citep{Belouadi2023AutomaTikZTS}.

\begin{figure*}[!ht]
\includegraphics[width=\textwidth]{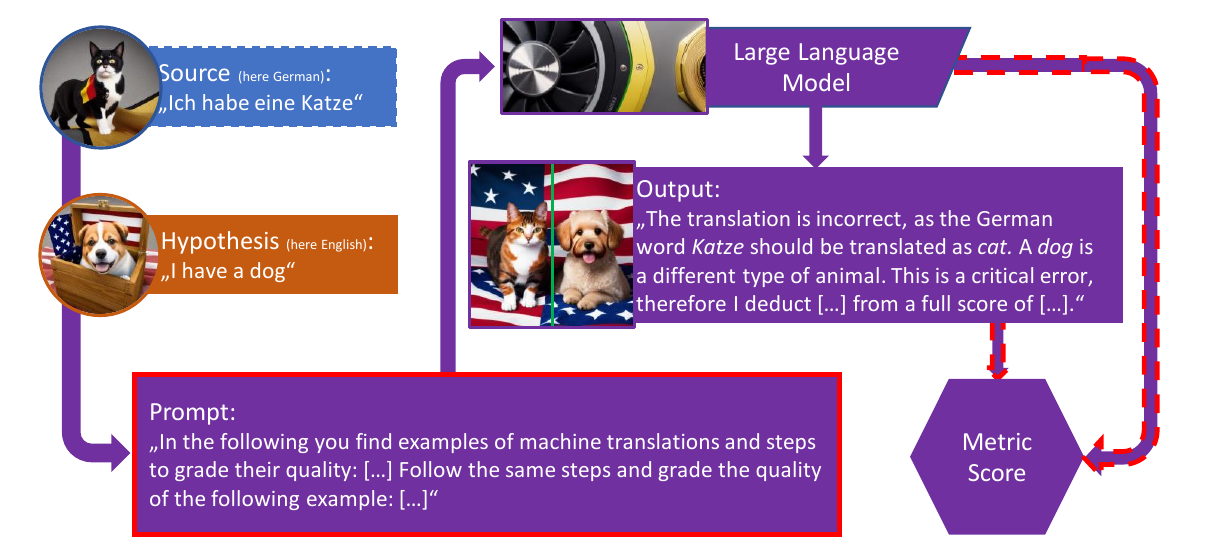}
\caption{Using a generative LLM as MT evaluation metric. In this example, the metric is reference-free, i.e., it grades the translated sentence based on its source sentence. The input sentences are wrapped into a prompt that is given to an LLM. The LLM generates an output and a final score could for example be extracted from this textual output or from other values involved in the process. The red borders indicate the focus of our shared task. Participants should evaluate the best prompts and the best approaches to \cll{extract} scores from model output.}
\label{shared_task_overview}
\end{figure*}

The immense research interest has also triggered the exploration of numerous approaches that leverage generative large language models (LLMs) as \textit{evaluation metrics} \citep{kocmi-federmann-2023-large, liu2023geval, fu2023gptscore, xu2023instructscore, fernandes2023devil} 
for natural language generation (NLG) tasks like machine translation (MT) and summarization. 
Recent LLM based approaches differ, for example, in their prompting strategies, e.g., in the way that natural language instructions are used to trigger the LLM to compute metric scores. For example, GEMBA \citep{kocmi-federmann-2023-large} uses zero-shot prompting to directly predict scores or quality labels in the output. In contrast, AutoMQM \citep{fernandes2023devil} instructs LLMs to predict fine-grained error labels and uses these to compute the final scores. \cl{These works have 
\se{initiated the idea} 
of prompting for NLG evaluation, but an exhaustive exploration of approaches remains} 
\se{an open problem.} 
Further, many approaches leverage closed source LLMs while much fewer use open source LLMs. Those approaches relying on 
open source LLMs put a large focus on acquiring training data \citep[e.g.] []{xu2023instructscore} and fine-tune models to specific tasks. Given this typical focus on fine-tuning and motivated by promising work on prompting techniques\footnote{Various websites track the development of prompting techniques (see Appendix \ref{promptingGuides}).} \citep[e.g.][]{wei2022chain, yao2023tree, wang2023selfconsistency, zhou2023large}, we notice a research gap in the thorough examination of \emph{prompting and 
\se{metric score extraction} 
in the domain of NLG metrics}, especially for \textbf{open source} generative LLMs \cll{(here, metric score extraction refers to the process of constructing the metric scores from internal parameters or the output of a model).}

The Eval4NLP 2023 shared tasks  
aims to fill this gap by disallowing participants to fine-tune models and by restricting model usage to a fixed list of LLMs (see Figure \ref{shared_task_overview}). Hence, participants may only vary 
how models are prompted, how scores are extracted, and 
how models are used in combination. \cl{To make the task more inclusive, we consider large and small(er) LLMs in two separate tracks. This is different from shared tasks without model restriction, where the largest models often perform best, for example, the WMT metrics shared task \citep[e.g.][]{freitag-etal-2022-results}.}

The goal of the shared task is to design evaluation metrics for MT and summarization, which we select as sub-tasks of NLG, while adhering to the model restrictions. Our contributions are the following: 
\begin{itemize}
    \item We design a novel, restricted evaluation setting that allows to focus on \emph{prompting and score extraction} in building evaluation metrics. This might aid inexpensive development of new metrics without fine-tuning or could benefit the selection of metric architectures with fine-tuning. 
    \item We collect a novel dataset from Wikipedia articles 
    created past 
    15.07.2023 with the goal of minimizing the use of data that has been used to pre-train LLaMA2 \citep{touvron2023llama} 
    released on 17.07.2023. This is 
    because 
    some of the allowed models are 
    fine-tuned versions of LLaMA2. 
    \item We organized a CodaLab \citep{codalab_competitions_JMLR} / Codabench \citep{codabench} competition where participants could submit their system scores in a dev and test phase. The dev phase has received 44 participant registrations, of which 9 teams have submitted contributions to the test phase leaderboard and system papers. This paper summarizes their approaches and findings and presents their final ranking. \cll{We find that the best performing submissions are on par with or surpassing metrics like Comet-Kiwi-XXL \citep{rei2023scaling} and GEMBA \citep{kocmi-federmann-2023-large} (that are not restricted by the shared task settings) on our test set. This is an interesting finding, as the submissions require less parameters and did not use fine-tuned models.}
    \item In line with the Eval4NLP 2021 shared task \citep{fomicheva-etal-2021-eval4nlp}, we consider the \emph{explainability} of the designed metrics. The generative nature of LLMs allows to return natural language or formatted explanations of 
    its output. While these explanations are not necessarily faithful, they also offer value if they are plausible \citep{Leiter2023TowardsEE} or might support the generation process itself \citep{wei2022chain}. 
\end{itemize}

Our paper is structured into \se{seven} sections. \S\ref{sec:related} gives an overview of how our shared task is related to other competitions \se{and presents related work on evaluation metrics}. 
\S\ref{sec:setup} describes the competition setup and \S\ref{sec:data} describes the datasets and annotation process for the test phase. In \S\ref{sec:approaches}, we highlight the approaches 
tested by the participants, especially those
for the test set submissions. \S\ref{sec:results} presents the final scores of the participants on the test set and further analyses. Finally, \S\ref{sec:conclusion} discusses future work and \se{concludes}. 
\section{Related Work} \label{sec:related}
In this 
\se{section}, 
we describe other work that is related to our shared task. 
\se{Specifically,} 
we give a brief overview of evaluation metrics, highlight 
recent development\se{s} on metrics 
based on generative LLMs and describe related shared tasks.

\begin{figure*}[!ht]
\includegraphics[width=\textwidth]{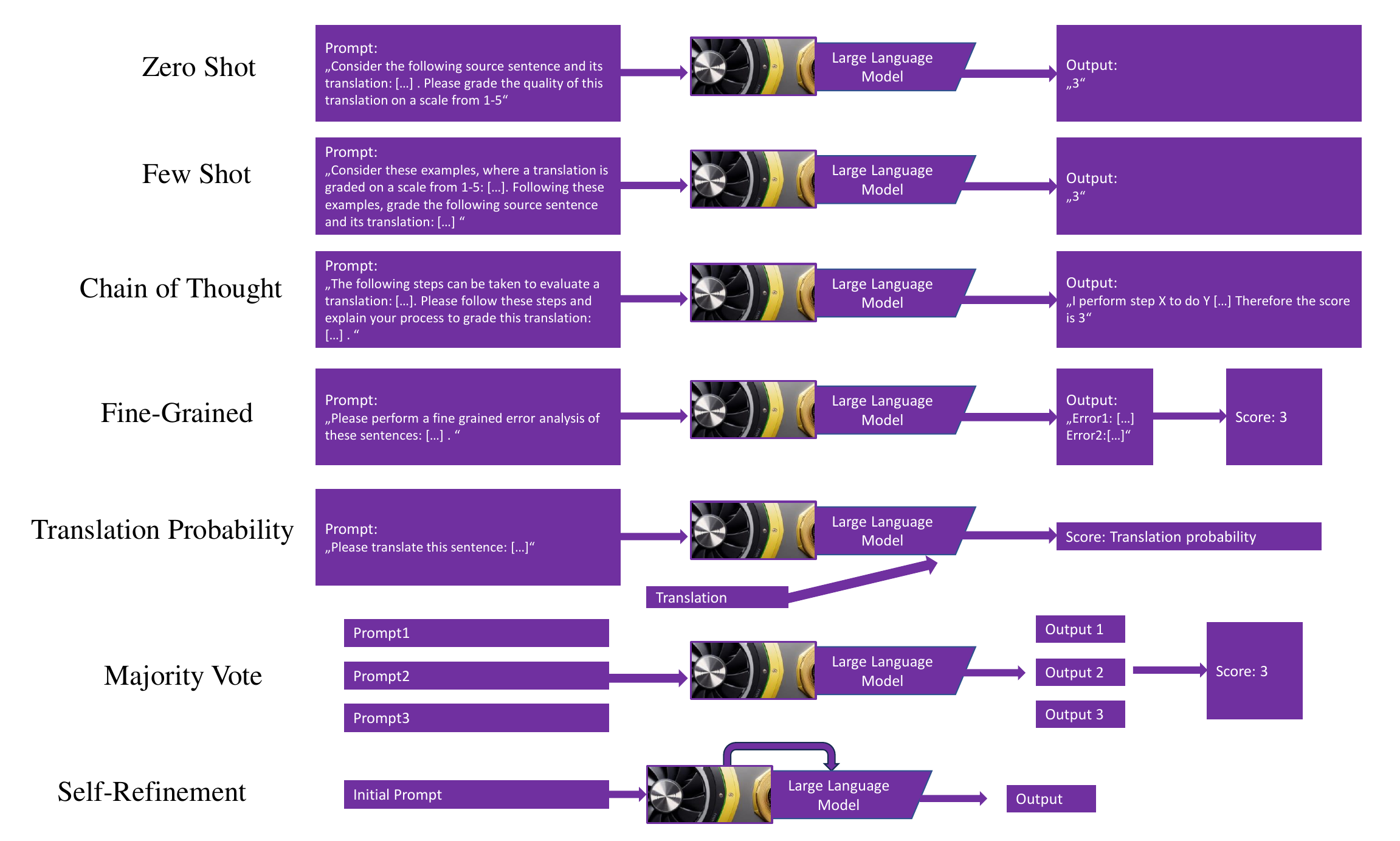}
\caption{Schematic overview of possible approaches to compute scores from generative LLMs. Zero-shot approaches do not present examples in the prompt, while few-shot approaches present them. Chain-of-though\se{t} \citep{wei2022chain} approaches trigger the LLM to generate an explanation of its process before returning the final score. Fine-grained approaches, e.g.\ \citet{fernandes2023devil}, first construct a detailed error analysis and then construct a final score from them. Translation probability approaches, e.g.\ \citet{fu2023gptscore}, use the probability of generating a paraphrase as a translation. In a majority vote approach\se{,} the results from multiple prompts could be combined. Self-refinement approaches could 
\se{query} 
a model multiple times to refine its output. Often, approaches can be combined- For example, chain-of-thought prompting can work with few shot prompting.}
\label{possible_approaches}
\end{figure*}

\paragraph{NLG evaluation metrics}
Evaluation of NLG systems is necessary \se{in order} to \se{be able to} compare \se{and rank} 
\se{different models or outputs}.  
Manual/human evaluation is expensive, time consuming and often infeasible for larger datasets. Hence, automatic metrics are constructed. 
Metrics that use 
\se{human} 
references are called \textit{reference-based}, while metrics that evaluate the generation quality based on the source text are called \textit{reference-free} (in MT also Quality Estimation, QE).
Many early metrics like BLEU \citep{papineni-etal-2002-bleu} and ROUGE \citep{lin-2004-rouge} measure the lexical overlap between the generation and a human written reference.

\se{Such metrics} 
\se{are limited} 
in their ability to capture semantics of generated text \citep[e.g.][]{reiter-2018-structured}. 
Newer metrics are usually based on language models 
able to embed the meanings of tokens \citep[e.g.][]{Zhang2020BERTScore, zhao-etal-2019-moverscore, sellam-etal-2020-bleurt, rei-etal-2020-comet}. These 
achieve strong\se{(er)} correlations to human judgments of generation quality \citep[e.g.][]{freitag-etal-2022-results}. Embedding based metrics have also enabled reference-free evaluation. This has the added benefit of no longer needing human reference generations and therefore enables further use cases, such as checking generation quality on the fly \citep[e.g.][]{zerva-etal-2022-findings}, training with metrics as supervision signal \citep[e.g.][]{wu-etal-2018-study} and using metrics during decoding \citep{fernandes-etal-2022-quality}. However, the usage of \se{(large)} black-box systems in the evaluation process also poses new challenges, e.g., regarding metric \textbf{efficiency} \citep{kamal-eddine-etal-2022-frugalscore,rei-etal-2022-searching,Grunwald2022CanWD}, \textbf{explainability} \citep{kaster-etal-2021-global,rei-etal-2023-inside,xu2023instructscore,guerreiro2023xcomet,Leiter2023TowardsEE}, and \textbf{robustness} \citep{chen_menli:2023,he-etal-2023-blind,vu-etal-2022-layer}.
\cll{Our work is also concerned with metric explainability and efficiency, because we (1) offer a small model participation track, (2) forbid fine-tuning, (3) allow only open source models and (4) perform a short experiment on the plausibility of explanations.}

Surveys on NLG metrics are presented by \citep[e.g.][]{celikyilmaz2021evaluation, 10.1145/3485766}.

\paragraph{Generation-based evaluation metrics} 
Generation-based evaluation metrics have \cll{become a fundamental part of} the current \cll{metric landscape},
beginning with PRISM \citep{thompson-post-2020-automatic} and BARTScore \citep{NEURIPS2021_e4d2b6e6}. 
These two metrics use the \textbf{generation probability} of paraphrases or translations as \cll{mechanism to extract} metric scores. Newer work that follows the same principle with more high-performing 
LLMs, \cll{for example GPTScore} \citep{fu2023gptscore}, has shown improved correlations.
Another branch 
of generation-based metrics has originated with recent GPT models and shows that models can directly perform the task of grading machine generated text from in-context task descriptions \citep[e.g.][]{kocmi-federmann-2023-large, chiang-lee-2023-large, fu2023gptscore, xu2023instructscore, yang2023knowledgeprompted, lu2023error}. We will refer to these metrics as \textbf{output-based}. Here, the rating is usually returned directly in the generated output text or constructed from it. \cl{Another branch of these models employs generative LLMs for ranking between better and worse generations \citep{zheng2023judging,shen2023large,ji2023exploring}.}

This recent surge of approaches has motivated our shared task. During the runtime of the shared task, other state-of-the-art approaches have been published \citep[e.g.][]{fernandes2023devil}. The systems submitted to our competition are different from most generation-based metrics in thoroughly exploring the usage of fixed recent open source LLMs since ChatGPT without 
fine-tuning.

\paragraph{Evaluation shared tasks} 
Our shared task is also related to other shared tasks that consider the evaluation of evaluation metrics for NLG, especially for MT and summarization. For MT, the established WMT workshop comprises multiple shared tasks on MT evaluation. Especially, the \textit{WMT metrics shared task} \citep[e.g.][]{mathur-etal-2020-results, freitag-etal-2021-results, freitag-etal-2022-results} and the \textit{WMT shared task on quality estimation} \citep[e.g.][]{specia-etal-2020-findings-wmt, specia-etal-2021-findings, zerva-etal-2022-findings} are related to ours. The main track of the \textit{WMT metrics shared task} considers the system- and segment-level evaluation quality of MT metrics ---  that is, how well can metrics reflect the quality of whole MT systems or single segment translations. Recent years also put a focus on evaluating the robustness of metrics towards certain linguistic phenomena. The main track of the \textit{WMT metrics shared task} consists of a reference-based evaluation, i.e., metrics compare the machine translation to human-written reference translations. Recent editions also contain a track for reference-free evaluation, where submitted metrics should directly compare the machine translation to its source text. Since 2021, the \textit{WMT metrics shared task} has acquired its test data using the fine-grained MQM evaluation scheme \citep{lommel-2014, freitag-etal-2021-experts} that has been shown to be more accurate than crowd-sourced direct assessment annotations. The \textit{WMT shared task on quality estimation} sets its main focus on the reference-free evaluation of machine translations. In recent years, their test sets are also annotated with MQM. Additionally, the quality estimation workshop has, for example, conducted tasks on word-level error prediction and span-level error severity prediction. 

Like the WMT QE shared task, our task is the reference-free evaluation of machine translations. The biggest difference of our shared task is that we fix the allowed models. That means, participants may only use models from a list we provide to them. Hence, participants have to focus on a thorough exploration of prompting and score extraction rather than fine-tuning and dataset creation. A second difference is that we include summarization as a subtask. As a third difference, our shared task has a subtrack to evaluate explanations that are created as a byproduct of scoring with generative LLM's for plausibility. This last point offers parallels to the Eval4NLP 2021 shared task \citep{fomicheva-etal-2021-eval4nlp} and its successor subtask at the WMT 2022 shared task \citep{zerva-etal-2022-findings} on quality estimation. These tasks treated human word-level error annotations as explanations of translation quality and evaluated their correlations to manual annotations. In our subtask, we allow for any kind of explanation. Background information on explainability for MT metrics can be found in \citet{Leiter2023TowardsEE}.

\paragraph{Prompting}
\cll{The main goal of our shared task is to explore the \textit{prompting} of LLMs as explainable metrics, i.e., to explore which natural language inputs trigger LLMs to perform as metrics for NLG evaluation. Besides the approaches used for generation-based metrics described above, many more prompting strategies have been developed (see Appendix \ref{promptingGuides}). Here, we give a non-exhaustive overview of prompting techniques (Figure \ref{possible_approaches} shows examples of prompting approaches applied to evaluation metrics). Generally inputs to the model could either be texts that should be completed by LLMs or they could be instructions for instruction-tuned LLMs \citep [e.g.][]{NEURIPS2022_b1efde53}.
\textit{Zero-shot prompting} simply provides a task description as model input, without presenting any examples \citep{wei2022finetuned}. In contrast, \textit{few-shot prompting} provides examples of correct solutions, which the model should generalize on for a new problem. \textit{Chain-of-thought (CoT) prompting} \citep{wei2022chain} triggers the model to output an explanation of computation steps before returning a result. CoT is setup either by providing similar explanation steps in the input or by prompting the model to produce explanations steps itself. Recent work finds, however, that CoT explanations may not necessarily be faithful \citep{turpin2023language}. CoT can be boosted with \textit{self-consistency} \citep{wang2023selfconsistency} (or majority vote) where multiple explanations are sampled and the most consistent one is chosen. Another prompting paradigm is \textit{tree-of-thought (ToT)}, where explanations are generated step-wise (termed \textit{thoughts}) and are used to construct a tree of most likely explanation paths based on which better solutions can be found.}
\cll{Other works consider the automatic construction of prompts. For example \citet{zhou2023large} propose \textit{automated prompt engineer (APE)}, a method to optimize prompts for LLMs. Also, \citet{NEURIPS2020_6b493230} propose \textit{retrieval augmented generation (RAG)} to incorporate external knowledge bases into prompt generation.}
\section{Shared Task Setup} \label{sec:setup}
As described in \S\ref{sec:introduction}, the goal of our shared task is to leverage generative LLMs as (explainable) metrics for MT and summarization.\footnote{We treat MT and summarization as separate tracks.} Thereby, participants are not allowed to fine-tune their models and only certain models are allowed. \cll{This leads to a setting in which participants mainly explore \textit{prompting strategies} and \textit{score extraction}. With \textit{score extraction}, we refer to the way in which the final metric score is constructed from an LLM. In \S\ref{sec:related}, we describe that metrics based on recent generative LLMs are either based on generation probabilities or directly on decoded textual output. Other options for score extraction could be embedding based, similar to BERTScore \citep{Zhang2020BERTScore}, or based on attention.} 

Figure \ref{shared_task_overview} shows the general setup of using generative LLMs as metrics, illustrated with an example from MT. The figure shows that final scores could be constructed from the generated model output or from other variables involved in the inference process. Specifically, recent works on prompting and metrics offer a wide range of possibilities to influence score construction even without fine-tuning. Some of them are shown in Figure \ref{possible_approaches}.

\paragraph{LLM sizes} We organize two tracks based on the model sizes. Models 
smaller than 25B parameters are considered as \textbf{small}, and models bigger than 25B parameters as \textbf{large}. Table \ref{tab:models} gives an overview of the allowed models. We mainly choose these models based on their good average performance on the Huggingface Open LLM Leaderboard.\footnote{\url{https://huggingface.co/spaces/HuggingFaceH4/open_llm_leaderboard}} For Platypus2, Guanaco and WizardLM, we use 4-bit quantized versions with GPTQ \citep{frantar2023gptq} to lower the system requirements to run them. 
Of these models, only the Guanaco model was explicitly fine-tuned with multilingual data. The models Wizard, Nous and Guanaco \cll{(fine-tunes of LLaMA \citep{touvron2023llama})} %\todo{SE: note the missing space}
were allowed for use from the start of the competition, while the other 3 models \cll{(fine-tunes of LLaMA2 \citep{touvron2023llama2})} %\todo{SE: missing space}
were added to the list 20 days later. In another track, we explore the explanatory value of explanations 
created as a byproduct of the scoring process (see \S\ref{sec:explanation_results}). 

\paragraph{Phases} Our shared task was conducted in two phases. 
First, we hosted a dev phase on CodaLab\footnote{\url{https://codalab.lisn.upsaclay.fr/competitions/15072}} \citep{codalab_competitions_JMLR} from 07.08.23 to 30.09.23. In this phase, participants were developing their approaches and could already evaluate their scores on a leaderboard. While the standing in the dev phase does not influence the ranking of the shared task, the phase aided the creation of a competitive atmosphere, acted as an advertisement for the competition and allowed us to gauge the number of interested participants. The main part of the competition was the test phase 
conducted from 26.09.23 to 01.10.23. Due to performance problems and unforeseen issues with extending the competition setup on CodaLab, the test phase was 
migrated to 
its successor Codabench\footnote{\url{https://www.codabench.org/competitions/1359/}} \citep{codabench}. Submissions to the dev phase and test phase both had to contain at least a file with newline separated scores that grade each sample of our datasets. The test phase additionally required to enter a team name, to indicate the track for each submission and to provide additional files with (1) a short system description, (2) newline separated prompts for each input, and (3) optionally newline separated explanations.

\begin{table*}
    \centering
    \begin{tabular}{l|l|l} \toprule
        \textbf{Mode} & \textbf{Release Date} & \textbf{Track}\\ \midrule
        Platypus2-70B-Instruct-GPTQ \citep{lee2023platypus}  & 11.08.23 & Large\\
        Guanaco-65B-GPTQ \citep{dettmers2023qlora}  & 25.05.23 & Large\\
        WizardLM-13B-V1.1-GPTQ \citep{xu2023wizardlm} & 07.07.23 & Small\\
        Nous-Hermes-13b  & 03.06.23 & Small\\
        OpenOrca-Platypus2-13B \citep{hunterlee2023orcaplaty1, mukherjee2023orca}  & 11.08.23 & Small\\
        orca\_mini\_v3\_7b \citep{orcaminiv37b, mukherjee2023orca}  & 07.08.23 & Small\\
        \bottomrule
    \end{tabular}
    \caption{Generative LLMs whose usage was allowed in the Eval4NLP 2023 shared task. We list the huggingface urls in Appendix \ref{urls}.}
    \label{tab:models}
\end{table*}

We describe the shared task datasets in \S\ref{sec:data}.

\section{Datasets} \label{sec:data}  
During the dev phase of our shared task, we provided participants with a train and a dev set. For the test phase, we further \cll{created} a test set.

\subsection{Train \& dev set} Our train and dev sets are constructed from two datasets. For MT, we select the en-de (English-German) and zh-en (Chinese-English) MQM partitions of the WMT 2022 metrics shared task \citep{freitag-etal-2022-results}. For summarization, we select SummEval \citep{fabbri-etal-2021-summeval}. We conduct our task in a reference-free setting, that is, we do not provide human written reference translations or summaries. 

Hence, we remove the references 
provided with WMT and SummEval. SummEval has separate scores for relevance, factuality, coherence and consistency for each sample. We construct a single score per example by averaging these separate scores.
Further changes to the original datasets include the split into train and dev partitions as well as shuffling. \cl{In the dev phase participants could experiment with generalizable (prompting) approaches.}

\subsection{Test set} \label{sec:test}
We collect a novel test set for the test phase of our shared task. It 
consists of 3 language pairs for MT: en-de (English-German), en-es (English-Spanish), en-zh (English-Chinese) and a summarization part. We only choose high-resource languages, as the allowed LLaMA(2)-based models have seen limited multilingual data during their pre-training and fine-tuning. Hence, high-resource languages can 
indicate 
an upper bound 
of what these models can achieve without further fine-tuning. 
To reduce the possibility that our chosen LLMs were trained on parts of the test set, we gather Wikipedia articles created after 15.07.23 as source texts.\footnote{Limitations of this approach are discussed in \S\ref{sec:limitations}.}

\paragraph{Annotator selection} For MT annotation, we hire one annotator per language pair: one 
\se{post-graduate} 
student who speaks Spanish as mother tongue with English certifications, one NLP Bachelor student, who is a native English speaker that lives in Germany since many years, 
and one data and discourse studies Master student, who is a native Chinese speaker who uses English on a daily basis.
For summarization annotation, we hire one NLP Bachelor student 
as well as a 
data and discourse studies Master student with a prior master in linguistics. Both annotators annotated the same data.
All annotators demonstrated their suitability for the role in initial test rounds with further applicants. The distribution of our final MT dataset is shown in Table \ref{SPPD}. \cl{The total annotation costs were ca.\ 5000€.}

\paragraph{Annotation tool} For MT and summarization, we perform fine-grained annotations. In MT, fine-grained MQM annotations have been shown to yield more reliable human annotations than other annotation schemes \citep{freitag-etal-2021-experts}. Also, the fine-grained annotations could be used later on to verify automatically generated explanations.\footnote{As we only received 2 submissions for the explainability track, we do not consider this in this report.} We use Google's Anthea\footnote{\url{https://github.com/google-research/google-research/tree/master/anthea}} as annotation tool, because of its support for \cll{such fine-grained} MQM annotations \citep{lommel-2014, freitag-etal-2021-experts}. As we mostly annotate single sentences for MT, we modify Anthea to provide context via a Wikipedia URL that can be consulted if annotators are unsure about a translation. For summarization, annotations were conducted in a modified version of Anthea with a new template (we show a screenshot of the UI in Appendix \ref{annotation_screenshot}). 

\paragraph{MT annotation} \cll{We construct the \textbf{MT} dataset from random sentences extracted from the Wikipedia source texts we collected.} Thereby, we select sentences with a minimum length of 110 characters, as tokenized by the NLTK sentence tokenizer\footnote{\url{https://www.nltk.org/api/nltk.tokenize.html}}. In a few cases, multiple sentences are concatenated due to missing spaces between dots. We obtain machine translations with 4 different translation models (see Table \ref{MTSModels}). Further, we use MQM as annotation scheme and conducted the annotation process in multiple batches to allow for corrections in subsequent batches. The batch sizes varied between 200 and 600 samples. \cl{For the first batch, we changed parts of the process during the annotation. }
Specifically, we had accidentally chosen an incorrect tokenization for the first few samples of the first batch.\footnote{For the test phase, 
we keep the annotations of the first batch, as small issues in source sentences should not invalidate the possibility of creating good translations; instead, we remove every sentence from the final dataset that has at least one major source error. \cl{We do this as major source errors might cause ambiguity in the annotation process. For example, if the source is unreadable, it is unclear which quality should be expected from the translation.}}
This may have led to coarser annotation and to ignoring some punctuation issues. We still use these samples, as punctuation errors only have a very small weight in MQM and a coarser annotation does not change the severity assigned to errors. Hence, we assume that the impact on the MQM scores is minimal. 
Another change between annotation versions is that the first batch contains unordered sentences, while in the second version, all translations of a single source follow each other (in a random order). This has majorly improved the annotation speed as annotators do not need to reread the source sentences anymore. Further, the annotators commented on difficult source texts in the first batch. Therefore, in the following batches, we pre-filter the Wikipedia source articles by their quality classes\footnote{\url{https://en.wikipedia.org/wiki/Wikipedia:Content_assessment}} and keep only c-class and better articles. Furthermore, we employ languagetool\footnote{\url{https://languagetool.org/de}} to filter for the grammatical correctness of the source sentences.

\begin{figure*}[!ht]
\begin{subfigure}[b]{0.42\textwidth}
            \centering
            \includegraphics[width=\textwidth]{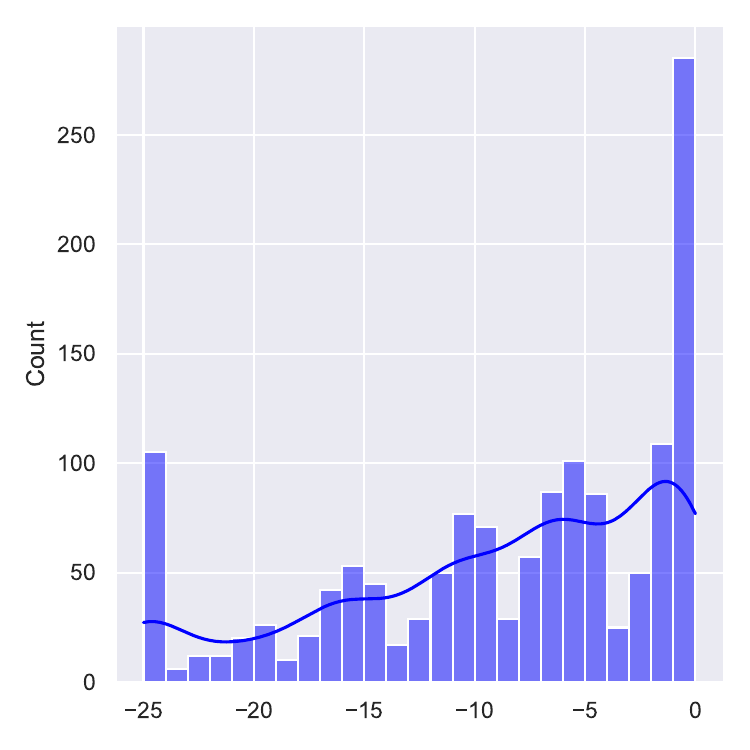}
            \caption[]
            {en-de}    
            \label{sd1}
        \end{subfigure}
        \hfill
        \begin{subfigure}[b]{0.42\textwidth}  
            \centering 
            \includegraphics[width=\textwidth]{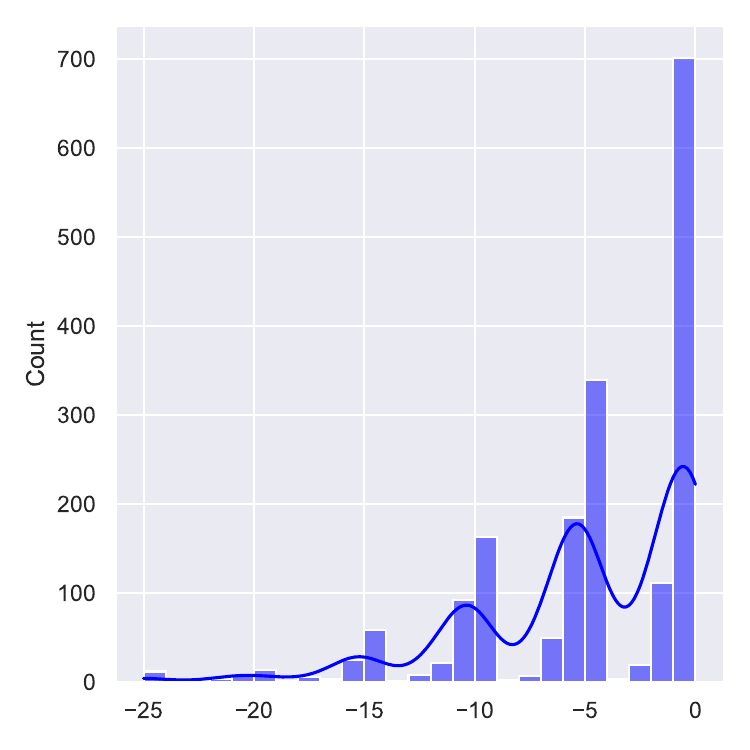}
            \caption[]
            {en-es}    
            \label{sd2}
        \end{subfigure}
        \vskip\baselineskip
        \begin{subfigure}[b]{0.42\textwidth}   
            \centering 
            \includegraphics[width=\textwidth]{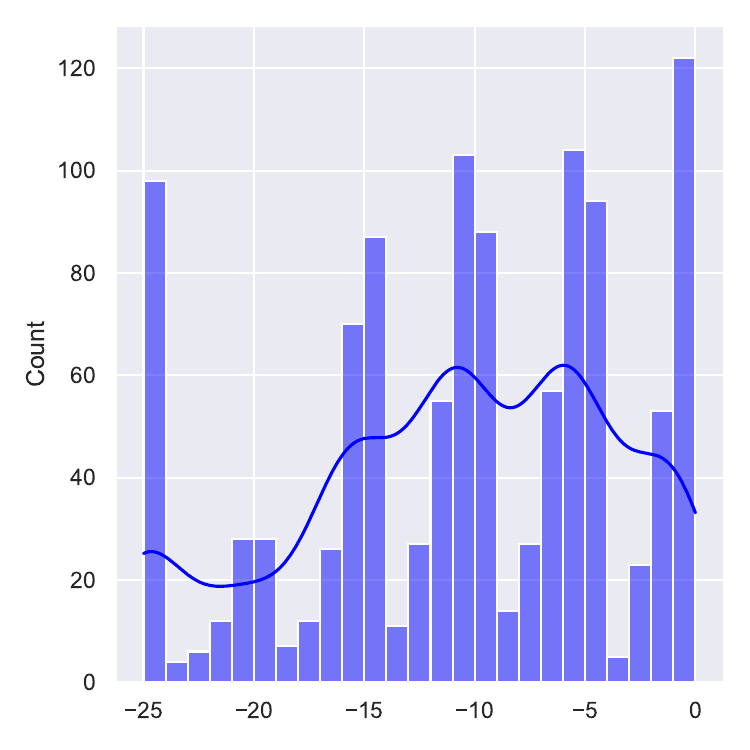}
            \caption[]
            {en-zh}    
            \label{sd3}
        \end{subfigure}
        \hfill
        \begin{subfigure}[b]{0.42\textwidth}   
            \centering 
            \includegraphics[width=\textwidth]{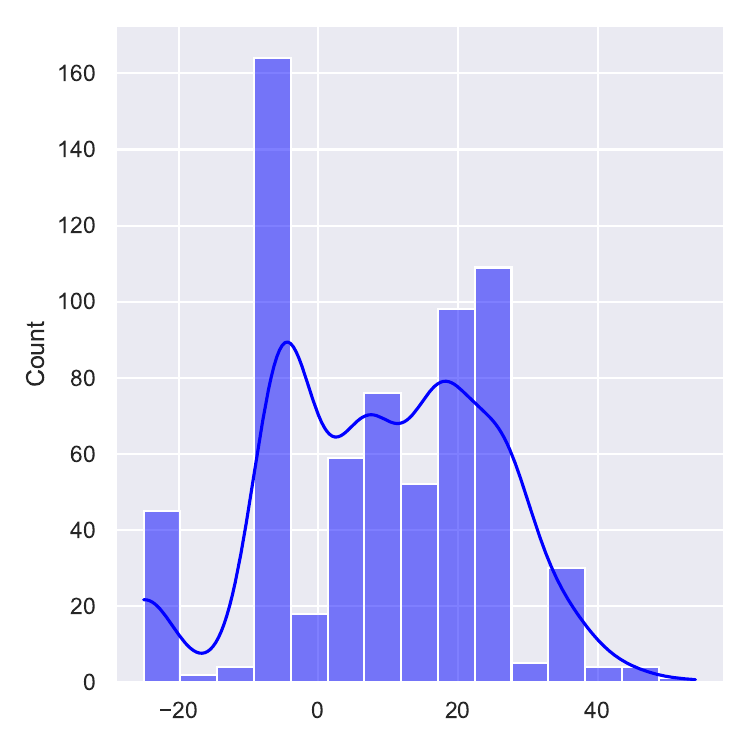}
            \caption[]
            {summarization}    
            \label{sd4}
        \end{subfigure}
\caption{\cll{MQM and summarization score} distributions of our datasets. \cll{These scores are constructed with heuristics from human annotations. The annotation process is described in \S\ref{sec:test}.} For MT, the scores range from -25 to 0. For summarization, they range from -25 to 54 (positive scores are possible because our heuristic rates relevant facts positively). The blue line indicates the kernel density estimation describing the distribution.}
\label{score_distribution}
\end{figure*}

\paragraph{\cll{MT annotator agreement}} To verify the quality of the dataset, members of our team who are native speakers of the respective target languages have annotated small subsets of 30-50 samples of the datasets. Table \ref{agreements} shows the agreement 
on these subsets, \se{which are acceptable, except for en-es}. \cl{For en-es, either the MT models 
performed better, 
the annotator might have 
\se{missed} 
some errors or they might have
\se{annotated} 
them less strictly, as suggested by Figure \ref{score_distribution}.}

\begin{table*}
    \centering
    \begin{tabular}{l|l}
        \toprule
        \textbf{MT Models} & \textbf{Summarization Models} \\ \midrule
        mbart50\_en2m \citep{JMLR:v22:20-1307} & sshleifer/distilbart-cnn-12-6 \citep{shleifer2020pretrained}\\
        mbart50\_m2m \citep{JMLR:v22:20-1307} & facebook/bart-large-cnn \citep{lewis-etal-2020-bart}\\
        m2m\_100\_418M \citep{tang-etal-2021-multilingual} & google/bigbird-pegasus-large-bigpatent \citep{NEURIPS2020_c8512d14}\\
        m2m\_100\_1.2B \citep{tang-etal-2021-multilingual} & facebook/bart-large-xsum \citep{lewis-etal-2020-bart}\\
        & mT5\_multilingual\_XLSum \citep{hasan-etal-2021-xl}\\
        \bottomrule
    \end{tabular}
    \caption{An overview of the translation and summarization models we have used to created our datasets. We list the urls in Appendix \ref{urls}.}
    \label{MTSModels}
\end{table*}

\begin{table}
    \centering
    \begin{tabular}{l|lll}
    \toprule
        \textbf{Type} & \textbf{Train} & \textbf{Dev} & \textbf{Test} \\ \midrule
        en-de & 11046 & 7364 & 1425 \\
        en-es & - & - & 1834\\
        en-zh & - & - & 1161 (1297)\\
        zh-en & 15750 & 10500 & -\\
        summarization & 320 & 1280 & 671 (825) \\
        \bottomrule
    \end{tabular}
    \caption{Number of samples in our datasets. In the case of the brackets, we filtered out potentially malformed examples after the test phase was conducted.}
    \label{SPPD}
\end{table}

\begin{table}
    \centering
    \begin{tabular}{cc}
    \toprule
        \textbf{Type} & \textbf{Agreement} \\ \midrule
        en-de & 0.458 \\
        en-es & 0.239\\
        en-zh & 0.480\\
        summarization & 0.625 (0.316/0.654) \\ \bottomrule
    \end{tabular}
    \caption{Kendall tau as agreement between annotators. For MT, the agreement was calculated on 30-50 samples. For summarization, it was calculated on 373 examples. \cll{The first value in brackets is the agreement on the MQM component of our heuristic and the second value is the agreement on the relevance/factuality component of our heuristic.}}
    \label{agreements}
\end{table}

\paragraph{Summarization annotation -- Concept}

To create a dataset that \cll{contains fine-grained and explainable human judgments of summary quality}, we perform a fine-grained annotation of three quality aspects defined by \citet{Dang2005OverviewOD,fabbri-etal-2021-summeval}: (1) \textit{factuality}, (2) \textit{relevance}, and (3) \textit{readability} (where readability includes the properties of coherence and fluency used by \citet{fabbri-etal-2021-summeval}). Factuality captures whether all facts in the summary correctly represent the source, relevance describes how relevant the summary is with respect to the source and readability includes properties such as the text being fluent, free from grammatical errors and easy to read. 
We note that readability is covered to a large degree by MT MQM annotation guidelines and build on them. We change \cll{these guidelines} by removing the category for \textit{adequacy} and adding \textit{coherence}. Based on \citet{Dang2005OverviewOD}, we create the MQM category for coherence with the following sub-categories: \textit{referential clarity}, \textit{redundancy}, \textit{structure}, and \textit{meaning}. The meaning category refers to cases where the summary changes the meaning of the source text without hallucinating, e.g., by concatenating facts in the wrong order.

\paragraph{Summarization annotation -- Factuality \& Relevance} 
While we base \textit{readability} on a variant of MQM, we annotate the \textit{factuality} and \textit{relevance} of summaries in a different way. 

One common approach to determine these two properties is 
the \textit{pyramid method} \citep{nenkova-passonneau-2004-evaluating}. Here, small atomic facts of many human written references are collected and ordered in a pyramid, based on their occurrence count. With this pyramid, it can be checked whether (1) facts in a summary are correct and (2) how relevant these facts are (facts at the top are more relevant then those at the bottom). 
Instead of the pyramid method, \cl{we apply a more resource efficient approach, where} we use a reference-free approach for annotating the summaries' relevance and factuality. Inspired by \citet{liu-etal-2023-revisiting}, who manually split the source text into atomic facts, we leverage the NLTK sentence tokenizer to split the source text into enumerated sentences. In some cases, sentences were not split correctly. In sentences of the final test set, we have corrected them manually. We treat each sentence as a single fact.\footnote{Splitting each sentence into more granular facts, might further improve the fine-grained score composition but would require more effort in determining distinct facts.} Next, we annotate the relevance of each of these facts, i.e., how likely would the annotator use the fact in a given sentence if they should write a summary 
themselves. Then, we annotate which source sentence is reflected in which part of the summary. By doing so, we can weigh the relevance of each fact that appears in the summary. Finally, we annotate each fact not represented in the original source text as a hallucination. 

\begin{figure*}[!ht]
\centering
\includegraphics[width=0.8\textwidth]{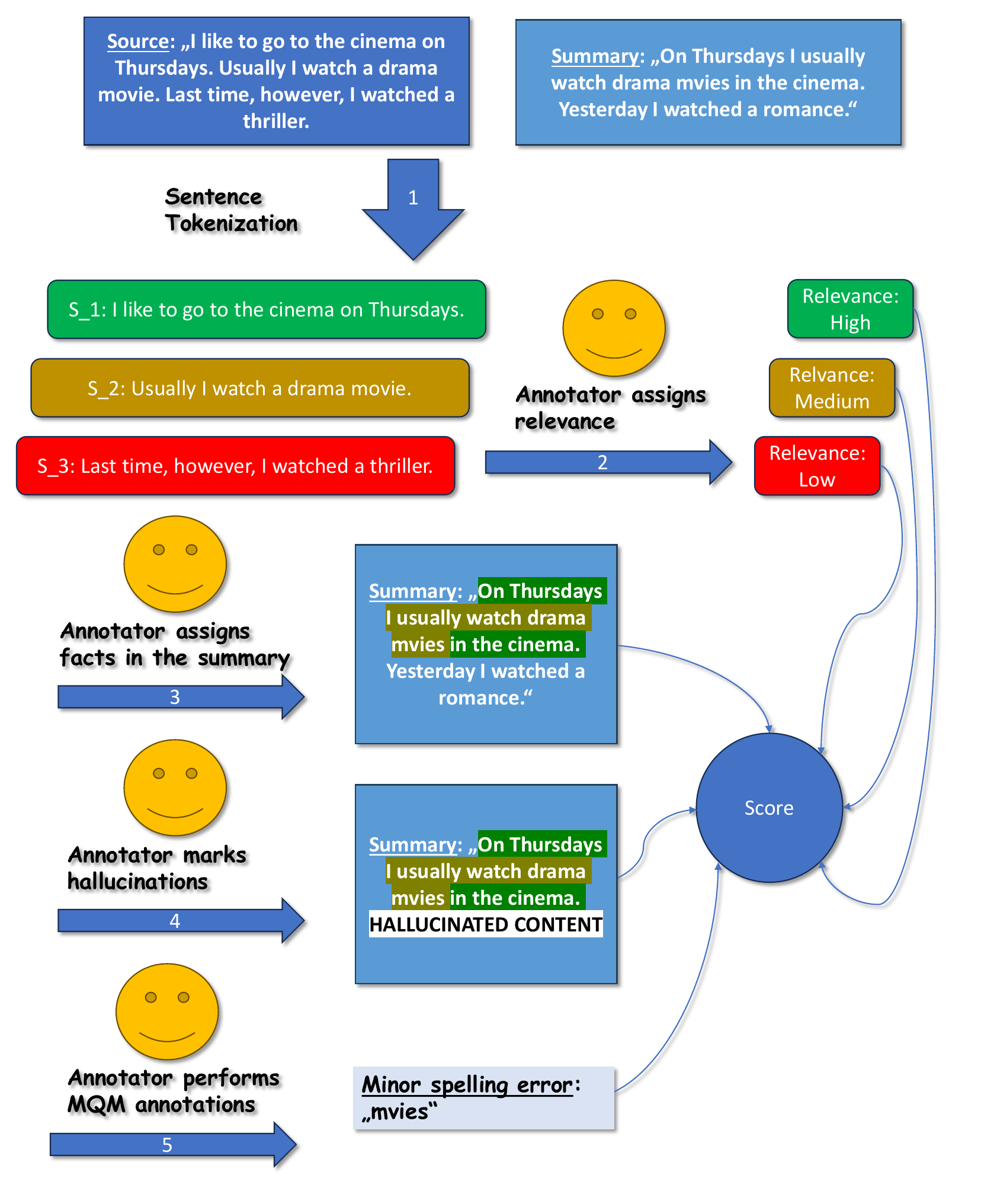}
\caption{An example of the summarization annotation process: \cll{(1) we automatically perform sentence tokenization of the source text, (2) annotators assign each sentence a relevance from low to high, (3) annotators match facts in the summary with sentences in the source text, (4) annotators mark hallucinated content in the summary and (5) annotators perform MQM error annotations of the summary. Based on steps 1, 2 and 3, we rate the relevance and factuality of the generated summary. Based on 4, we rate the amount of hallucinations, which also contributes to factuality. Finally, based on 5, we rate the readability of the summary.}}
\label{Summarization_set_example}
\end{figure*}

\paragraph{Summarization annotation- Score} Based on the annotations for \textit{factuality}, \textit{relevance} and \textit{readability}, we build a heuristic that is negative for bad summaries and positive for good summaries. The equation is shown in Figure \ref{sum_eq}.
\begin{figure*}
\begin{equation}
\sum_{i\in \text{Facts\;in\;Summary}} \alpha*\text{Relevance}(i) + \beta*\frac{|\text{Hallucinated\;Characters}|}{|\text{Characters\;in\;the\;summary}|} + \gamma * \text{MQM}
\end{equation}
\caption{A heuristic for fine-grained reference-free evaluation of summaries. We set $\alpha=3$, $\beta=5$ and $\gamma=1$.}
\label{sum_eq}
\end{figure*}
Here, $\alpha$, $\beta$ and $\gamma$ can be chosen to determine the influence of each sub-score for relevance, hallucinations and readability, respectively. There are many design choices regarding the weighting of each component and different normalization approaches. We find that these generally only have a small impact on the final ranking of our shared task (see Appendix \ref{heurisitc_impact}). Longer summaries can contain more facts and would hence receive higher scores in this heuristic. We address this issue by generating summaries of similar lengths using max token settings.
The example in Figure \ref{Summarization_set_example} shows this annotation process.

\paragraph{Summarization annotation - Process} We select random sections from Wikipedia that have a length of 150 to 800 tokens as measured by the tokenizer of \textit{bart-large-cnn}. The summarization models we use are listed in Table \ref{MTSModels}. Like with MT, we annotated in several batches. 

After the first batch, as for MT, we took measures to improve the source quality and ordered the sources to allow for faster annotations. After a check on the annotation quality, some misunderstandings of the annotation classes were uncovered and discussed. In the final evaluation, we drop all examples 
labeled before this discussion, such that we keep a total of 671 samples.
Further, one annotator showed a larger annotation speed \cl{and a more consistent understanding of the task}. In the test set, we use the annotations of this annotator. \cll{We do not use the average here. When we would use the average, we would either have to use the smaller number of samples that was annotated by both annotators or we would need to mix average scores and the samples from the faster annotator.
%\todo{SE: `samples average scores' misses a genetive ', I think. Maybe "mix averages scores with the samples from the faster annotator" is easier, though?}
}

Table \ref{agreements} shows the agreement between the annotators. It is high for relevance and factuality annotations and lower for the MQM part. 

%\todo{SE: Table 8 should slightly smaller regarding width}

\paragraph{\cll{Score distributions}}
Figure \ref{score_distribution} shows the distributions of scores constructed from human annotations in our test set, \cll{i.e., the scores that are used as ground truth in our evaluation}. We can see that all language pairs exhibit a pattern of centering around values divisible by 5. This makes sense, as MQM weighs major errors with 5 points. Also, in \textit{en-es}, samples have generally received a higher score; i.e., fewer major errors were annotated. Finally, our summarization dataset, which uses a combined annotation scheme does not show this pattern.

\subsection{Evaluation}
Following earlier WMT tasks on segment-level evaluation, we compute Kendall's tau correlation \citep{10.1093/biomet/33.3.239} to compare the system generated scores to human scores.
We further report 
Spearman and Pearson correlations.\footnote{For these evaluations of correlations, we use the implementations of the python scipy library: \url{https://scipy.org/}} Future work could explore if the usage of other and possibly more suited variants of Kendall, as suggested by \citet{deutsch2023ties}, might affect the rankings of our competition.
\section{Shared Task Approaches} \label{sec:approaches}
The test phase of our shared task received submissions from 12 different teams, 9 of which submitted system papers. Here, we summarize the approaches of these 9 systems and announce their final standings. Table \ref{approaches} gives an overview of the participating teams and of the tracks they are participating in.\footnote{While the first and last authors of \citet{NLLG} are members of the NLLG group, we did not share any internal details that would have given them an advantage. They developed their approach independently.} This table can be used as a mapping for the scores reported in \S\ref{sec:results}.

\begin{table*}
    \centering
    \begin{tabular}{l|c|c}
    \toprule
       \textbf{Team} & \textbf{Authors} & \textbf{Tracks} \\ \midrule
        DSBA & \cite{DSBA} & S, L, SU \\
        iML & \cite{iML} & S, L, SU \\
        IUST\_NLP\_Lab & \cite{IUSTNLPLab} & S, SU, E \\
        HIT-MI\&T Lab & \cite{HITMI} & S, MT \\
        Kotonya et. al.& \cite{CompetitionEntrants} & S, SU \\
        LTRC & \cite{LTRC}& S, MT, SU, E \\
        NLLG & \cite{NLLG}& L, MT, SU \\
        Pradhan/Todi & \cite{Beginners} & S, SU \\
        TaiwanSenior & \cite{TaiwanSenior} & S, MT \\
        \bottomrule
    \end{tabular}
    \caption{Overview of shared task submissions. The letters are abbreviations for the following tracks: S(mall model track), L(arge model track), M(achine)T(ranslation track), SU(mmarization track), E(xplainability track). }
    \label{approaches}
\end{table*}

We divide the approaches taken by the participants \cll{based on their method to score extraction} into \textit{probability-based}, \textit{output-based} and \textit{agent-based}.\footnote{View \S\ref{sec:related} for the distinction of probability-based and output-based.} Besides their final approaches, the participants have explored a large number of possible variations, \cll{both of which we summarize}. 
We \se{also} introduce the baseline approaches we compare the participants \se{to}.

\paragraph{Probability-based}
Probability-based approaches calculate how likely a paraphrase or translation of an input is generated with an LLM. Probability based approaches are explored by HIT-MI\&T Lab and Pradhan/Todi. 
HIT-MI\&T Lab define 10 different prompts to translate a source sentence with an LLM. They combine this approach with \cll{retrieval augmented generation and few-shot prompting by} demonstrating samples in the input prompt 
selected by (among others) SBERT \citep{reimers-gurevych-2019-sentence}.  Further, they use ensembles to recombine the scores of multiple prompts and models. Pradhan/Todi use the probability-based approach with own prompts and prompts 
designed by the authors of GPTScore \citep{fu2023gptscore}. They also explore ensembles across 4 different prompts.

\paragraph{Output-based}
All submitted papers explore the direct usage of an LLM's natural language output as score. HIT-MI\&T Lab test the same sample selection and ensembling strategies described above with 4 different prompts in an output-based setting. NLLG follow a similar approach to HIT-MI\&T Lab and retrieve demonstration examples by finding similar examples with LABSE \citep{feng-etal-2022-language} embeddings in an output-based setting. Pradhan/Todi try one approach in which they present a prompt that triggers the prediction of a single score and one approach that triggers the model to first rate summary qualities for consistency, coherence, fluency and relevancy. Then they aggregate these scores in 3 different ways. LTRC quantize Orcamini themselves to run an even smaller model  (which is close to 
violating the allowed settings of the shared task). 
They provide a detailed explanation to their model that triggers it to produce fine-grained scores and a combined score in the same output. DSBA choose rating guidelines from related work ---  
concretely, the human guidelines (HG) for SummEval, the machine guidelines for G-Eval \citep{liu2023geval} and evaluation steps generated by GPT4 \citep{gpt4}. They test various adaptations to this prompt, explore the usage of examples in the prompt and the usage of coarse-grained vs.\ fine-grained and aggregated scores. On the test set, they add a shortcut for very bad summarizations and employ bucketing for their scores. iML explores evaluating 6 different criteria over all model combinations. Kotonya et. al. explore 8 prompt types: 3 base prompts and their extensions with chain-of-though \citep{wei2022chain}, zero-shot and few-shot settings. IUST\_NLP\_Lab explores various zero-shot and few-shot settings with Orcamini.
Finally, IUSTNLPLab and LTRC generate explanations as an additional request to their model. 

\paragraph{Agent-based}
While they also use an output-based setup, we place TaiwanSenior in a separate group. They define 4 characters that should be played by a model and a list of 10 properties. For example they define ``Internet Troll'' as a critical character or ``Teacher'' as more knowledgeable character, with the intention that different viewpoints can help to judge generation quality better.
Then, they evaluate the combined 40 settings and use XGBoost \citep{Chen_2016} to combine their scores. While they did not add their top submissions to the final leaderboard they present their reasonably good final scores in their paper. 

\paragraph{Baselines}
As baselines, we apply the widely used metrics BERTScore (with XLMR-large embeddings) \citep{Zhang2020BERTScore}, SBERT \citep{reimers-gurevych-2019-sentence} cosine-similarity (with XLMR-large embeddings), SUPERT \citep{gao-etal-2020-supert} (with 3 settings), GEMBA \citep{kocmi-federmann-2023-large} and Comet-Kiwi-XXL \citep{rei2023scaling}. These baselines use models that are not allowed as part of the shared task. \cll{Therefore, we refer to them as external baselines and abbreviate them as follows in \S\ref{sec:results}: \textit{ex-BaselineBERTScore}, \textit{ex-BaselineSBERT}, \textit{ex-BaselineSupertMP(net2)}, \textit{ex-BaselineSupertF(ull)}, \textit{ex-BaselineSupert5}, \textit{ex-BaselineGEMBA} and \textit{ex-BaselineComet(KiwiXXL)}.}
Further, we include one baseline for every allowed model that uses the DA score prompt of GEMBA \citep{kocmi-federmann-2023-large} (with a slight modification for summarization). Following the order in table \ref{tab:models}, in \S\ref{sec:results} we abbreviate these baselines with \textit{baselinePlaty\_lg}, \textit{baselineGuanaco\_lg}, \textit{baselineWizard}, \textit{baselineNous}, \textit{baselineOrcaPlaty} and \textit{baselineOrcaMini}. The models are further specified in Appendix \ref{model_details}.
\section{Results and Analysis} \label{sec:results}

\begin{table*}
\begin{tabular}{l|rrr|rrr|rrr}
\toprule
\multicolumn{1}{c}{}& \multicolumn{3}{c}{Kendall} & \multicolumn{3}{c}{Pearson} & \multicolumn{3}{c}{Spearman}\\
Team & de & zh & es & de & zh & es & de & zh & es \\
\midrule
\textbf{HIT-MI\&T Lab} & \textbf{0.491} & \textbf{0.375} & \textbf{0.417} & \textbf{0.655} & \textbf{0.528} & \textbf{0.453} & \textbf{0.656} & \textbf{0.511} & \textbf{0.553} \\
\textit{ex-BaselineGEMBA} & \textbf{0.492} & \textbf{0.384} & \textbf{0.409} & 0.506 & 0.356 & 0.251 & \textbf{0.625} & \textbf{0.496} & 0.512 \\
\textit{ex-BaselineComet} & 0.421 & \textbf{0.345} & 0.288 & 0.562 & 0.443 & 0.331 & 0.583 & \textbf{0.484} & 0.403 \\
\textit{ex-BaselineBertscore} & 0.239 & 0.174 & 0.221 & 0.344 & 0.236 & 0.179 & 0.344 & 0.252 & 0.312 \\
\textit{ex-BaselineSBERT} & 0.209 & 0.167 & 0.226 & 0.246 & 0.210 & 0.081 & 0.304 & 0.242 & 0.320 \\
\textbf{LTRC} & 0.194 & 0.144 & 0.112 & 0.232 & 0.133 & 0.031 & 0.233 & 0.173 & 0.132 \\
\textit{baselineNous} & 0.189 & 0.011 & 0.112 & 0.183 & 0.044 & 0.045 & 0.230 & 0.013 & 0.136 \\
\textit{baselineOrcaPlaty} & 0.189 & 0.011 & 0.112 & 0.183 & 0.044 & 0.045 & 0.230 & 0.013 & 0.136 \\
seanstilwell & 0.120 & NaN & NaN & 0.164 & NaN & NaN & 0.152 & NaN & NaN \\
\textit{baselineOrcaMini} & 0.073 & 0.188 & 0.065 & 0.030 & 0.102 & 0.009 & 0.088 & 0.225 & 0.077 \\
\textit{baselineWizard} & 0.101 & 0.065 & 0.079 & 0.047 & 0.057 & 0.026 & 0.121 & 0.077 & 0.093 \\
\textbf{TaiwanSenior} & 0.041 & NaN & NaN & -0.037 & NaN & NaN & 0.051 & NaN & NaN \\
\bottomrule
\end{tabular}
\caption{\cll{Results of the \textit{small} model track for MT, ordered by the mean of correlations (for columns without NaN). Each column shows the correlation of metric scores to MQM scores for English-X language pairs}. Results that are bolded are significantly better than non-bolded results, with $p\leq0.05$, as measured by a permute-both significance test \citep{deutsch-etal-2021-statistical}. Teams with paper submissions are bolded.
}

\label{small_mt}
\end{table*}

\begin{table*}
\begin{tabular}{l|rrr|rrr|rrr}
\toprule
\multicolumn{1}{c}{}& \multicolumn{3}{c}{Kendall} & \multicolumn{3}{c}{Pearson} & \multicolumn{3}{c}{Spearman}\\ 
Team & de & zh & es & de & zh & es & de & zh & es \\
\midrule
\textit{baselinePlaty\_lg} & \textbf{0.362} & \textbf{0.293} & \textbf{0.264} & \textbf{0.312} & \textbf{0.270} & 0.129 & \textbf{0.445} & \textbf{0.364} & \textbf{0.320} \\
\textit{baselineGuanaco\_lg} & \textbf{0.350} & 0.219 &\textbf{ 0.241} & \textbf{0.344} & 0.176 & 0.125 & \textbf{0.445} & 0.273 & \textbf{0.300} \\
\textbf{NLLG} & 0.245 & 0.139 & 0.179 & 0.257 & 0.196 & \textbf{0.155} & 0.335 & 0.190 & 0.238 \\
\textbf{kaiwalya\_large} & 0.174 & 0.113 & 0.125 & 0.161 & 0.141 & 0.052 & 0.209 & 0.138 & 0.147 \\
\bottomrule
\end{tabular}
\caption{Results of the \textit{large} model track for MT, ordered by the mean of correlations (for columns without NaN). Each column shows the correlation of metric scores to MQM scores for English-X language pairs. Results that are bolded are significantly better than non-bolded results, with $p\leq0.05$, as measured by a permute-both significance test \citep{deutsch-etal-2021-statistical}. Teams with paper submissions are bolded.} 
\label{large_mt}
\end{table*}

\begin{table}
\begin{tabular}{lrrr}
\toprule
Team & kd & ps & sp \\
\midrule
\textbf{DSBA} & \textbf{0.633} & \textbf{0.783} & \textbf{0.782} \\
\textbf{iML} & \textbf{0.615} & \textbf{0.763} & \textbf{0.772} \\
\textit{ex-BaselineBertscore} & 0.578 & 0.\textbf{771} & \textbf{0.765} \\
\textit{ex-BaselineSupertMP} & 0.554 & 0.736 & \textbf{0.747} \\
\textbf{IUST\_NLP\_Lab} & 0.573 & 0.722 & \textbf{0.722} \\
\textit{baselineOrcaMini} & 0.560 & 0.681 & \textbf{0.706} \\
\emph{ex-BaselineSupertF} & 0.516 & 0.686 & \textbf{0.706} \\
\textbf{Kotonya et. al.} & 0.546 & 0.680 & \textbf{0.682} \\
\textbf{LTRC} & 0.531 & 0.691 & \textbf{0.679} \\
\textit{baselineOrcaPlaty} & 0.552 & 0.666 & \textbf{0.674} \\
\textit{baselineNous} & 0.552 & 0.666 & \textbf{0.674} \\
\emph{ex-BaselineSupert5} & 0.492 & 0.654 & \textbf{0.678} \\
\textit{ex-BaselineSBERT} & 0.465 & 0.625 & \textbf{0.645} \\
\textit{baselineWizard} & 0.411 & 0.534 & 0.536 \\
\textbf{Pradhan/Todi} & 0.436 & 0.032 & 0.610 \\
Haaland & 0.221 & 0.514 & 0.280 \\
\bottomrule
\end{tabular}
\caption{\cll{Results of the \textit{small} model track for summarization, sorted by the mean of correlations.} \textit{kd} stands for Kendall, \textit{ps} stands for Pearson and \textit{sp} stands for Spearman. Results that are bolded are significantly better than non-bolded results, with $p\leq0.05$, as measured by a permute-both significance test \citep{deutsch-etal-2021-statistical}. Teams with paper submissions are bolded.}
\label{small_s}
\end{table}

\begin{table}
\begin{tabular}{lrrr}
\toprule
Team & kd & ps & sp \\
\midrule
\textbf{DSBA} & \textbf{0.603} & \textbf{0.756} & \textbf{0.766} \\
\textbf{iML} & \textbf{0.612} & \textbf{0.738} & \textbf{0.768} \\
\textit{baselinePlaty\_lg} & \textbf{0.600} & \textbf{0.740} & \textbf{0.753} \\
\textbf{NLLG} & 0.471 & 0.643 & 0.638 \\
\textit{baselineGuanaco\_lg} & 0.402 & 0.492 & 0.504 \\
\bottomrule
\end{tabular}
\caption{Results of the \textit{large} model track for summarization. \textit{kd} stands for Kendall, \textit{ps} stands for Pearson and \textit{sp} stands for Spearman. Results that are bolded are significantly better than non-bolded results, with $p\leq0.05$, as measured by a permute-both significance test \citep{deutsch-etal-2021-statistical}.  Teams with paper submissions are written in bold.}
\label{large_s}
\end{table}

In this section, we 
first report 
statistics of the shared task. 
Then, we 
present and discuss the final system ranking. Note that we include submissions of participants on the test set leaderboard that did not submit a system paper. However, we do not describe their approaches in \S\ref{approaches}. Lastly, we 
discuss the implications of these results on the development of generation-based metrics. 

\paragraph{Statistics} The dev phase on CodaLab has received 44 registrations, 13 of which have submitted their scores. In total, there have been 1048 submissions on the dev set suggesting that some participants might have optimized their method on the dev set. Especially, one participant submitted 417 submissions on the dev set. The test phase on Codabench has received 21 registrations and 248 submissions from 11 participants. We have restricted the number of allowed submissions per day to 10. \cl{Allowing a higher number would enable participants to optimize their approaches on the test set too much, 
\se{so} that the results would not reflect the generalization capability anymore. On the other hand, we wanted to give participants the option to try out multiple approaches they designed. Further, Codabench would sometimes fail to compute scores and still deduct one submission. Hence, ten submissions per day \cll{allow to accommodate such scenarios.}}
Two participants have used up a contingent of $\approx 50$ submissions. Of the 11 test phase participants, 9 have submitted a system paper. The first authors are from China, India (2), Korea, Taiwan, Canada, Iran, Germany and the United Kingdom.

\se{Thus,} 
many authors are from developing countries. Also, many authors are students. Hence, their resource availability was limited, leading many of them to opting for smaller models. 

\paragraph{Correlation with humans} Here, we present the results that the participants achieve on the test sets. A mapping between team names and authors can be found in Table \ref{approaches}. Table \ref{small_mt} shows the final ranking of the \textit{small} MT subtask. 
Compared to the other participants, HIT-MI\&T Lab leads by a large margin on all correlation measures. It even outperforms the recent ex-BaselineCometKiwiXXL significantly and is only matched by ex-BaselineGEMBA with GPT-4 (in our baselines). \cll{Notably, both of these models have many more parameters.} 
\cll{This ranking is surprising, as the scores they report on the dev set are still better than their baselines using the same models, but not \se{comparatively} strong as ex-BaselineCometKiwiXXL (see the \textit{discussion} paragraph in this section)}. The test set approach that HIT-MI\&T Lab report in their paper builds on ensembling probability-based scores from prompts to OpenOrca-Platypus. These prompts contain \se{between} 3 up to 5 example demonstrations via retrieval augmented generation.\footnote{In their paper they describe that they use the maximum number of examples. However, this number is capped to 5 by their implementation.}
Future work should explore whether their approach can uphold its strong performance across other datasets and settings. The ranking is then followed by various baseline models and team LTRC, which used their chain-of-thought prompting + fine-grained approach for en-de and zero shot prompting for the other two language pairs.

Table \ref{large_mt} shows the final ranking of the \textit{large} MT subtask. 
For this subtask, the baselines have not been beaten. \cll{Interestingly, NLLG who also use retrieval augmented generation, perform worse than HIT-MI\&T Lab, even though they use a larger model. We assume that this is mainly caused by the latter using a probability-based approach, while NLLG use an output based approach. Potentially the translation capability of the LLMs (with next-word prediction) is larger than their ability to rate multilingual text.}

Table \ref{small_s} shows the final ranking of the \textit{small} summarization subtask. DSBA and iML lead this track. In their final submission, DSBA uses zero-shot prompting in a fine-grained evaluation setting, where they create an ensemble over 3 different prompts for relevance, factuality and readability. On the other hand, the final submission of iML is using a single zero-shot prompt that asks the model to rate the syntax of an input summary with respect to its source. 

Table \ref{large_s} shows the final ranking of the \textit{large} summarization subtask. Again iML and DSBA perform on par using the same methods they used for the small models. 
Interestingly, for MT and summarization, the small models (Tables \ref{small_mt} and \ref{small_s}) have beaten the large models. One potential reason might be that the large models take much longer to run and therefore they could not be examined with the same care. Further, it is interesting that \textit{baselineOrcaMini} and IUST\_NLP\_Lab beat many other models with OrcaMini despite its parameter count being the lowest of the allowed models'. \cl{Generally, many teams opted for the usage of small models. Some teams only use the OrcaMini model, due to resource constraints. This highlights 
\se{how the usage of smaller models in metrics fosters inclusiveness.}
We show a further analysis of the impact of the summarization subcategories in Appendix \ref{sec:subcategories}.}

\paragraph{Performance}
The best performing approaches of the participants achieve a similar Kendall correlation as our team members when we were testing the inter-annotator agreement on a small subset of samples (see \S\ref{sec:setup}). This suggests that these approaches are already close to the performance of native speakers that did little training with the annotation process (as compared to our main annotators with a strong language background and more annotation experience on the task). \cll{This result is similar to the findings of \citet{freitag-etal-2021-experts}, where metrics outperformed crowd sourced DA annotation scores.}

This is an intriguing finding and highlights the potential of current open source models with and without fine-tuning. \cll{Also, many prompting approaches like tree-of-thoughts or self-refinement still remain to be explored, which extends this potential. }
Further, it shows that for closed source models like ChatGPT or GPT4, similar opportunities may exist and lead to new state-of-the-art metrics. The results also show that comparably small hardware can already be enough to create strong new metrics.

\paragraph{Explainability} \label{sec:explanation_results}
Only two participants, LTRC and IUSTNLPLab, have submitted entries with complementary explanations to the Coda\se{B}ench leaderboard. Both directly prompted the model to give reasoning for the model's decision. Thus, we perform the human experiment on explainability only on a small scale of 50 annotations for randomly selected samples of our summarization dataset. Two annotators of our team were presented with source, summary, MQM annotations (to help to identify problems), the scores of the participants and the explanations of the participants. They annotated which of two explanations they prefer. One annotator preferred explanations of one system (let us call it A) in 27 cases and explanations of the other in 23 cases. The other annotator preferred system A in 24 cases and the other system in 26 cases. In these annotations,  the annotators agree in 56\% of cases. 
These findings show that the annotators did not have a clear preference between the systems. 
Also, we notice that many explanations tend to be vague and return texts such as ``The summary has a good coherence and fluency''. In some cases, the explanations correctly describe problems. We show one example explanation of LTRC in
Table \ref{table:summExample}. Here, the explanation correctly captures the word repetition. 

\begin{table*}
\begin{tabularx}{\textwidth}{Bss}
\toprule
\textbf{Source} & \texttt{Summary}  & \texttt{Explanation}\\
\midrule
 In 1918, Wimble built a small ship at Hastings with help from a friend and sailed to the West Indies to seek his fortune after his family faced financial hardship. In 1922, he acquired land in the Bahamas which enabled him to begin trading with the English colonies in mainland North America. He also acquired land in North Carolina, which was formally granted to him by George Burrington's council on August 4, 1723. Wimble later moved to Boston, Massachusetts where he married Rebecca Waters, the daughter of a prominent local, on March 26, 1724. Their first son, James, was born on December 20, 1724. He owned land in the South End which he presumably operated as a distillery. While in Boston, he continued his business of moving trade goods between North Carolina and various British trade posts in the West Indies. This business enabled him to increase his land holdings in North Carolina and purchase a brigantine, which he named "Rebecca" after his wife. In 1932, Wimble lost his ship and all of its cargo to a hurricane after being forced by Governor Woodes Rogers of the Bahamas to use his ship to protect vessels and salt ponds in Rum Cay. Wimble was forced to sell a portion of his belongings, land, and slaves to cover the loss and began the process of trying to collect damages from Woodes Rogers' commandeering of his ship.\footnote{This paragraph is from \url{https://en.wikipedia.org/wiki/James_Wimble}}  & Wimble is the subject of this patent application. Wimble is the son of Wimble. Wimble is the son of Wimble. Wimble is the son of Wimble. Wimble is the son of Wimble. Wimble is the son of Wimble. Wimble is the son of Wimble. Wimble is the son of Wimble. Wimble is the son of Wimble. & Explanation: The summary is not relevant to the source text, as it contains no information about Wimble's life, family, or business ventures. Instead, it repeats the name "Wimble" multiple times, which is not a summary of the source text. \\
\bottomrule
\end{tabularx}
\caption{Explanation generated with the approach by LTRC. It correctly identifies the issue of the word \textit{Wimble} repeating often. }
\label{table:summExample}
\end{table*}

\paragraph{Discussion} \label{devanalysis}
\cll{The strong performance of the leading systems for MT and summarization is promising. For MT, HIT-MI\&T Lab achieve competitive performance to state-of-the-art metrics on the test set. Their metric exhibits a performance/memory-tradeoff. They use a smaller 13B parameter model, compared to the much larger Comet-Kiwi-XXL and (likely) GEMBA with GPT4. Due to this, they are more memory efficient. On the other hand, the probability based-approach requires one forward-pass for every hypothesis token. Further, they ensemble between 10 outputs, resulting in a computation heavy process. In contrast, Comet-Kiwi-XXL and GEMBA directly predict the output scores. 
iML and DSBA (for summarization) use prompt-based approaches that require only the score tokens to be generated, hence they are more efficient than probability-based approaches. On the other hand, they have more parameters than baselines such as SUPERT (with the models we have applied).}

\cll{Some of the participants also report their results on the dev sets we constructed from WMT22. %\todo{SE: need to say WMT22 dev set, I guess}
For MT, we compute the baselines that use allowed models and Comet-Kiwi-XXL on the dev-set. HIT-MI\&T Lab report correlations of $0.250$ (en-de) and $0.319$ (zh-en). The best baseline with allowed models, baselineGuanaco, achieves $0.260$ (en-de) and $0.311$ (zh-en). Comet-kiwi-XXL achieves $0.295$ (en-de) and $0.337$ (zh-en). While HIT-MI\&T Lab is still very good on the dev set and beats other small model baselines by a large margin, it does not outperform the large models. This could be caused by several reasons: (1) The final approach of HIT-MI\&T Lab might have been slightly different. (2) The dev set is built from WMT22 which uses data from various domains (e.g. news and social media) as a source, while we create our test set from Wikipedia. (3) We use 4 MT systems, while WMT22 is more diverse. (4) Our annotators have assigned more major errors per average than the annotators for WMT. This could be caused by (2) and (3). In Appendix \ref{dev_stats},  we show the distributions of the dev sets. Future work could verify the viability of their method on further datasets and with further models.}

\cll{For summarization, both iML and DSBA report a score of 0.45 on the dev set %\todo{SE: hyphen or no?}
using small models. This surpasses baselines we computed on the dev set. The closest one is \textit{baselinePlaty\_large}, with $0.44$. Tha means, summarization models on the dev set achieve a similar ranking to the test set. %\todo{SE: which yields a similar standing? Don't understand}
These similar scores raise the question whether there is some upper bound for each model that can be reached with different prompts.}

\section{Conclusion} \label{sec:conclusion}
We  
summarize the shared task and discuss future work.

\paragraph{
\se{Summary \& Implications}} This work describes the 
Eval4NLP 2023 shared task on \textit{prompting LLMs as explainable metrics}. We have constructed a fine-grained dataset for MT and summarization evaluation, with a novel annotation scheme for the latter. Further, we have organized a competition following the novel restriction to specify allowed models and disallow fine-tuning in a MT and summarization evaluation setting. By running a small and a large model track, we have enabled participation for participants with fewer resources, leading to an inclusive shared task setting. 

The top scores of the participants highlight a number of interesting findings that we summarize here:
\begin{itemize}
    \item \textbf{Small Models}: The results on the test set show that the best solutions built on small models outperform those that are built on larger models. This is contradicting usual patterns and an interesting 
    \se{implication} for metric efficiency. 
    \item \textbf{Probability-based vs.\ Output-based}: The MT ranking is led by a probability-based method, while the summarization ranking is led by two output-based methods. 
    \cll{The allowed LLMs have seen little multilingual training data. Therefore, their understanding of other languages than English could be smaller than their capability of translation, hence favoring probability-based methods. Also, the winner for MT is using retrieval augmented generation, which might infuse more multilingual capabilities into the model.} 
    \item \textbf{Simplicity helps}: 
    Many baseline systems achieved high ranks, despite using a simple prompting approach. Participants often report that demonstrating examples reduced their performance. The two best performing approaches for summarization use zero-shot prompting. Hence, lean metrics are easier to design and can still be very powerful. The best ranked systems, however, explore more intricate prompts \cll{and ensembles}.
\end{itemize}

The contributions of our participants highlight once more how current LLMs can achieve state-of-the-art performance, even without any task-specific fine-tuning. 

\paragraph{Future Work} We have considered high resource languages for the MT task. Future work could evaluate low resource 
languages, especially once more generative LLMs are released that are trained across a wide range of languages. 
Also, 
\se{prospectively one might} 
encourage and set rewards for pipeline-based solutions. 
In other words, currently most approaches of the shared task are based on single prompts or probability outputs; instead many interesting approaches like tree\se{-}of\se{-}thoughts \citep{yao2023tree} explore pipelines in which the output is generated iteratively or in parallel. 
Future work might also create larger or more diverse datasets for our evaluation scheme. Another point is that our current work only contains a small analysis of explainability that remained indecisive on the explanation quality between two participants. This could be extended in future work.

\section*{Acknowledgements}
We thank our participants for the active contribution and discussion. Further, we thank our annotators for their effort in creating our test sets. 
Christoph Leiter is financed by the BMBF project ``Metrics4NLG''. Steffen Eger is financed by DFG Heisenberg grant EG 375/5–1.

\section*{Limitations} \label{sec:limitations}
One potential limitation of our work lies in the usage of data from Wikipedia after 15.07. \cll{Although the chosen articles were indeed picked after July 15th, it is important to 
note that some of the content may have been duplicated from elsewhere, a few texts were automatically translated from existing entries in other languages, and there is
the possibility that some of the content was computer-generated.} 
Another \cll{limitation} of our work lies in the low agreements between our team member and our Spanish annotator for the small test conducted. \cll{Hence, the quality of our Spanish dataset is less proven and} evaluation on Spanish might be less accurate than the other two language pairs. Due to time restrictions, we could not do further evaluations.
Still, we believe that our Spanish annotator was capable in their language and thorough with their analysis of the samples. 
As another limitation, pre-filtering with language tool and later on sorting out severe source errors might miss out on more subtle errors causing problems in the test set.

\bibliography{anthology,custom}

\appendix

\section{Prompting Guides} \label{promptingGuides}
Here we present a list of resource collections on prompting:
\begin{itemize}
    \item \url{https://www.promptingguide.ai/}
    \item \url{https://github.com/promptslab/Awesome-Prompt-Engineering}
    \item \url{https://github.com/DukeLuo/awesome-awesome-prompts}
    \item \url{https://github.com/snwfdhmp/awesome-gpt-prompt-engineering}
    \item \url{https://github.com/dqxiu/ICL_PaperList}
    \item \url{https://github.com/EgoAlpha/prompt-in-context-learning}
\end{itemize}

\section{Impact of the summarization heuristic}
\label{heurisitc_impact}
Here, we consider the impact of using alternative heuristics for summarization, by studying their effect on the ranking of summarization systems. The results for Equation \ref{sum_eq2} are shown in Table \ref{small_s2}. The results for Equation \ref{sum_eq3} are shown in Table \ref{small_s3}. We can see that the top rankings remain the same.

\begin{table}
\begin{tabular}{lrrr}
\toprule
team & s\_kd & s\_ps & s\_sp \\
\midrule
DSBA & 0.623 & 0.675 & 0.772 \\
iML & 0.602 & 0.642 & 0.757 \\
IUST\_NLP\_Lab & 0.566 & 0.678 & 0.712 \\
bertscore & 0.546 & 0.711 & 0.729 \\
baselineOrcaMini & 0.545 & 0.640 & 0.684 \\
Kotonya et.al. & 0.543 & 0.745 & 0.675 \\
baselineOrcaPlaty & 0.527 & 0.589 & 0.650 \\
baselineNous & 0.527 & 0.589 & 0.650 \\
LTRC & 0.522 & 0.655 & 0.666 \\
baselineSBERT & 0.438 & 0.524 & 0.611 \\
Pradhan/Todi & 0.424 & 0.030 & 0.594 \\
baselineWizard & 0.408 & 0.489 & 0.531 \\
Haaland & 0.265 & 0.732 & 0.332 \\
cometXXL & -0.009 & 0.091 & -0.015 \\
baselineSUPERT & -0.028 & -0.040 & -0.040 \\
\bottomrule
\end{tabular}
\caption{Results of the \textit{small} model track for summarization with Equation \ref{sum_eq2}.}
\label{small_s2}
\end{table}

\begin{table}
\begin{tabular}{lrrr}
\toprule
team & s\_kd & s\_ps & s\_sp \\
\midrule
DSBA & 0.551 & 0.490 & 0.695 \\
iML & 0.533 & 0.454 & 0.687 \\
ISUT\_NLP\_Lab & 0.512 & 0.546 & 0.649 \\
bertscore & 0.497 & 0.569 & 0.663 \\
baselineOrcaMini & 0.485 & 0.517 & 0.612 \\
Kotonya et.al. & 0.480 & 0.690 & 0.604 \\
LTRC & 0.476 & 0.534 & 0.609 \\
baselineOrcaPlaty & 0.462 & 0.446 & 0.581 \\
baselineNous & 0.462 & 0.446 & 0.581 \\
Pradhan/Todi & 0.422 & 0.023 & 0.591 \\
baselineSBERT & 0.384 & 0.371 & 0.539 \\
baselineWizard & 0.361 & 0.381 & 0.478 \\
Haaland & 0.295 & 0.800 & 0.368 \\
cometXXL & 0.015 & 0.159 & 0.021 \\
baselineSUPERT & 0.003 & -0.018 & 0.004 \\
\bottomrule
\end{tabular}
\caption{Results of the \textit{small} model track for summarization with Equation \ref{sum_eq3}.}
\label{small_s3}
\end{table}

\section{Impact of subcategories} \label{sec:subcategories}
We also study the impact of subcategories on the final ranking of summarization. That means, we calculate the ranking with each of $\alpha$, $\beta$, $\gamma$ set to 1, while the others are 0. The results are shown in Tables \ref{small_s4}, \ref{small_s5} and \ref{small_s6}. Intriguingly, when only the MQM score is evaluated, the model by \textit{Haaland} has the highest correlation. However, they did not submit a system description or a system paper. Further, all baselines in this setting perform relatively weak. The best baseline is comet, potentially as it has been trained on MQM scores. The results for relevance and hallucinations are rather unsurprising with one time \textit{DSBA} being the winning team and the other time \textit{iML}.

\begin{table}
\begin{tabular}{lrrr}
\toprule
team & s\_kd & s\_ps & s\_sp \\
\midrule
Haaland & 0.334 & 0.796 & 0.379 \\
DSBA & 0.172 & 0.401 & 0.210 \\
Kotonya et. al. & 0.166 & 0.642 & 0.200 \\
IUST\_NLP\_LAB & 0.164 & 0.472 & 0.200 \\
cometXXL & 0.163 & 0.184 & 0.215 \\
Pradhan/Todi & 0.158 & 0.022 & 0.205 \\
LTRC & 0.154 & 0.462 & 0.191 \\
iML & 0.146 & 0.362 & 0.174 \\
baselineWizard & 0.133 & 0.327 & 0.163 \\
baselineOrcaMini & 0.126 & 0.447 & 0.155 \\
baselineOrcaPlaty & 0.100 & 0.370 & 0.120 \\
baselineNous & 0.100 & 0.370 & 0.120 \\
bertscore & 0.097 & 0.481 & 0.130 \\
baselineSBERT & 0.071 & 0.293 & 0.094 \\
baselineSUPERT & 0.023 & -0.013 & 0.030 \\
\bottomrule
\end{tabular}
\caption{Results of the \textit{small} model track for summarization, when only predicting MQM.}
\label{small_s4}
\end{table}

\begin{table}
\begin{tabular}{lrrr}
\toprule
team & s\_kd & s\_ps & s\_sp \\
\midrule
DSBA & 0.600 & 0.730 & 0.727 \\
iML & 0.596 & 0.720 & 0.722 \\
bertscore & 0.562 & 0.687 & 0.724 \\
IUST\_NLP\_LAB & 0.553 & 0.637 & 0.677 \\
baselineOrcaMini & 0.549 & 0.595 & 0.669 \\
baselineOrcaPlaty & 0.536 & 0.606 & 0.638 \\
baselineNous & 0.536 & 0.606 & 0.638 \\
Kotonya et. al. & 0.522 & 0.525 & 0.634 \\
LTRC & 0.511 & 0.608 & 0.635 \\
baselineSBERT & 0.464 & 0.594 & 0.616 \\
Pradhan/Todi & 0.397 & 0.023 & 0.543 \\
baselineWizard & 0.393 & 0.479 & 0.491 \\
Haaland & 0.164 & 0.280 & 0.197 \\
baselineSUPERT & -0.041 & -0.059 & -0.056 \\
cometXXL & -0.065 & -0.083 & -0.092 \\
\bottomrule
\end{tabular}
\caption{Results of the \textit{small} model track for summarization, when only predicting relevance.}
\label{small_s5}
\end{table}

\begin{table}
\begin{tabular}{lrrr}
\toprule
team & s\_kd & s\_ps & s\_sp \\
\midrule
iML & 0.516 & 0.599 & 0.606 \\
bertscore & 0.471 & 0.480 & 0.595 \\
DSBA & 0.454 & 0.576 & 0.537 \\
baselineOrcaPlaty & 0.432 & 0.483 & 0.506 \\
baselineNous & 0.432 & 0.483 & 0.506 \\
Pradhan/Todi & 0.414 & 0.041 & 0.532 \\
baselineOrcaMini & 0.406 & 0.417 & 0.487 \\
baselineSBERT & 0.403 & 0.477 & 0.525 \\
IUST\_NLP\_LAB & 0.391 & 0.421 & 0.469 \\
LTRC & 0.353 & 0.382 & 0.429 \\
Kotonya et.al. & 0.348 & 0.220 & 0.417 \\
baselineWizard & 0.267 & 0.331 & 0.323 \\
baselineSUPERT & -0.031 & -0.043 & -0.041 \\
Haaland & -0.067 & -0.127 & -0.077 \\
cometXXL & -0.198 & -0.212 & -0.265 \\
\bottomrule
\end{tabular}
\caption{Results of the \textit{small} model track for summarization, when only predicting hallucinations.}
\label{small_s6}
\end{table}

\begin{figure*}
\begin{equation}
\sum_{i\in \text{Facts\;in\;Summary}} \alpha*\text{relevance}(i) + \beta*\frac{|\text{Hallucinated\;Characters}|}{|\text{Characters\;in\;the\;summary}|} + \gamma * \text{MQM}
\end{equation}
\caption{A heuristic for fine-grained reference-free evaluation of summaries. Alternatively, we set $\alpha=1$, $\beta=1$ and $\gamma=1$.}
\label{sum_eq2}
\end{figure*}

\begin{figure*}
\begin{equation}
\frac{\sum_{i\in \text{Facts\;in\;Summary}} \alpha*\text{relevance}(i)}{|\text{Facts\;in\;Source}|} + \beta*\frac{|\text{Hallucinated\;Characters}|}{|\text{Characters\;in\;the\;summary}|} + \gamma * \text{MQM}
\end{equation}
\caption{An alternative heuristic for fine-grained reference-free evaluation of summaries. We set $\alpha=1$, $\beta=1$ and $\gamma=1$. Further, we divide the relevance part by the number of facts in the source as normalization.}
\label{sum_eq3}
\end{figure*}

\section{Screenshot of the annotation interface}
\label{annotation_screenshot}
\begin{figure*}[!ht]
\includegraphics[width=\textwidth]{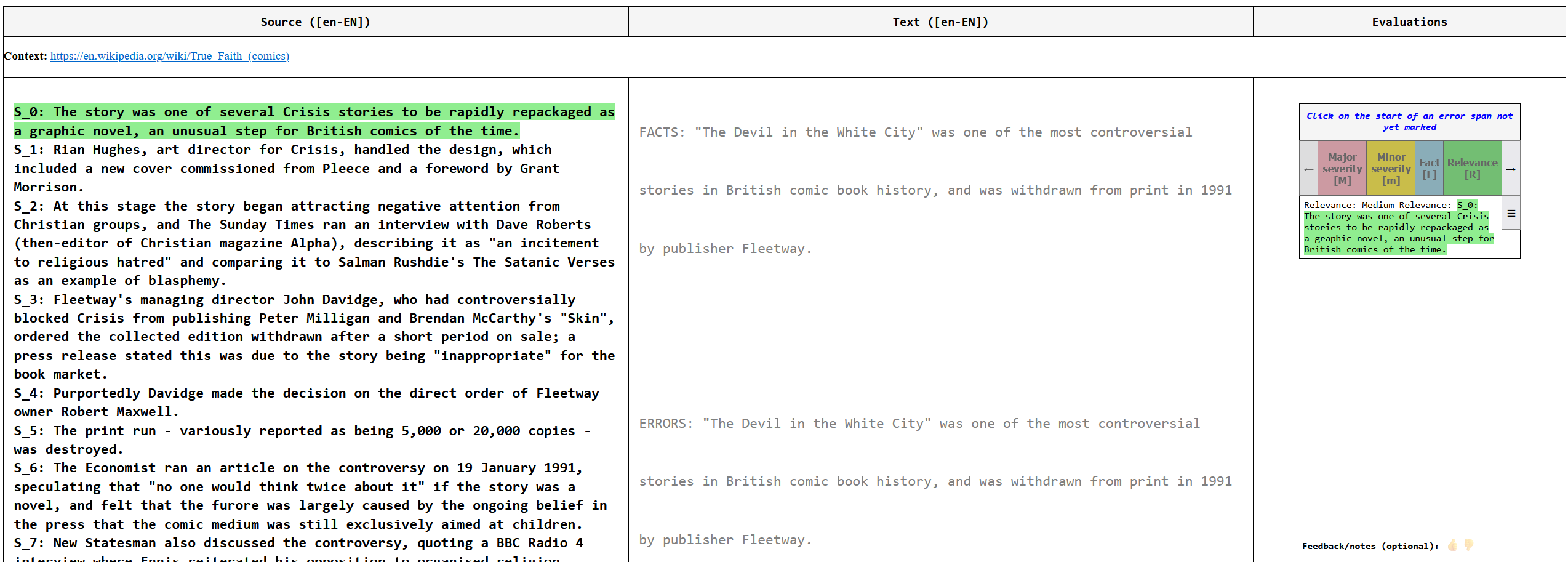}
\caption{The modified anthea annotation interface for summarization.}
\label{interface_anthea_summ}
\end{figure*}

Figure \ref{interface_anthea_summ} shows a screenshot of the Anthea annotation interface.

\section{Urls} \label{urls}
Here, we list the huggingface urls of the models we have allowed in the shared task:
\begin{itemize}
    \item Platypus2-70B-Instruct-GPTQ: \url{https://huggingface.co/TheBloke/Platypus2-70B-Instruct-GPTQ}
    \item guanaco-65B-GPTQ: \url{https://huggingface.co/TheBloke/guanaco-65B-GPTQ}
    \item WizardLM-13B-V1.1-GPTQ: \url{https://huggingface.co/TheBloke/WizardLM-13B-V1.1-GPTQ}
    \item Nous-Hermes-13b: \url{https://huggingface.co/NousResearch/Nous-Hermes-13b}
    \item OpenOrca-Platypus2-13B: \url{https://huggingface.co/Open-Orca/OpenOrca-Platypus2-13B}
    \item orca\_mini\_v3\_7b: \url{https://huggingface.co/pankajmathur/orca\_mini\_v3\_7b}
\end{itemize}

Further, here is a list of the urls we have used to translate/summarize our test set:
\begin{itemize}
    \item We use the implementations of the EasyNMT library by Nils Reimers for MT: \url{https://github.com/UKPLab/EasyNMT}
    \item distilbart-cnn-12-6: \url{https://huggingface.co/sshleifer/distilbart-cnn-12-6}
    \item bart-large-cnn: \url{https://huggingface.co/facebook/bart-large-cnn}
    \item bigbird-pegasus-large-bigpatent: \url{https://huggingface.co/google/bigbird-pegasus-large-bigpatent}
    \item bart-large-xsum \url{https://huggingface.co/facebook/bart-large-xsum}
    \item mT5\_multilingual\_XLSum \url{https://huggingface.co/csebuetnlp/mT5_multilingual_XLSum}
\end{itemize}

\section{Model Details}\label{model_details}
For SBert, we use embeddings of XLM-R to include multilinguality\footnote{\url{https://huggingface.co/sentence-transformers/stsb-xlm-r-multilingual}}. For SUPERT we report the standard metric using bert-large-nli-stsb-mean-tokens\footnote{\url{https://huggingface.co/sentence-transformers/bert-large-nli-stsb-mean-tokens}} with 5 and all source sentences as pseudo-references. Further, we upgrade SUPERT to use all-mpnet-base-v2\footnote{\url{https://huggingface.co/sentence-transformers/all-mpnet-base-v2}}, which improves its performance. For COMET, we use comet-kiwi-xxlm\footnote{\url{https://huggingface.co/Unbabel/wmt23-cometkiwi-da-xxl}}, which achieved strong results on reference-free evaluation. Fort GEMBA we use the GEMBA library\footnote{\url{https://github.com/MicrosoftTranslator/GEMBA}} and make small modifications to support GPT-4 requests. Finally, for BERTScore, we use xlm-roberta-large\footnote{\url{https://huggingface.co/xlm-roberta-large}}.

\section{Dev-Set Distribution}
\label{dev_stats} 
\begin{figure*}[!ht]
\begin{subfigure}[b]{0.42\textwidth}
            \centering
            \includegraphics[width=\textwidth]{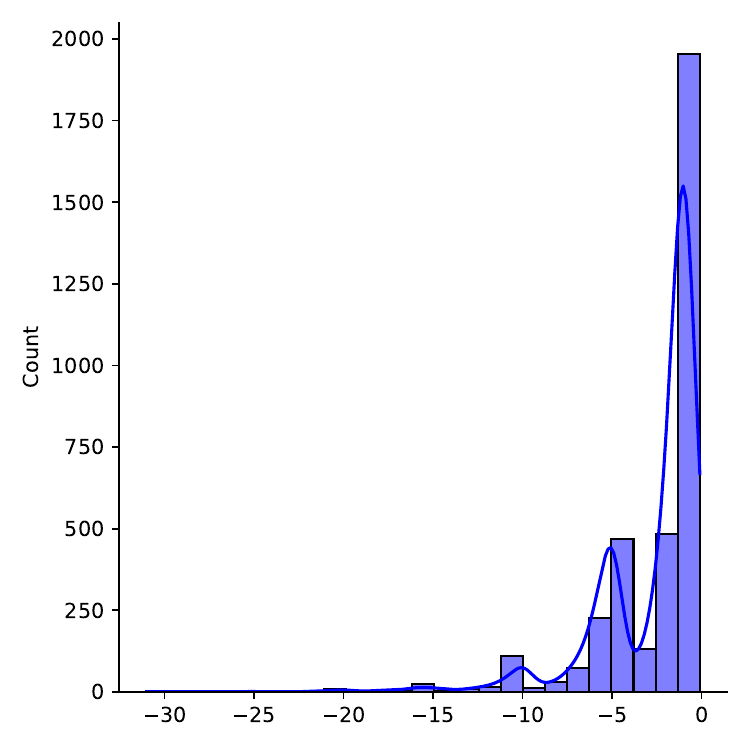}
            \caption[]%
            {en-de}    
            \label{sdd1}
        \end{subfigure}
        \hfill
        \begin{subfigure}[b]{0.42\textwidth}  
            \centering 
            \includegraphics[width=\textwidth]{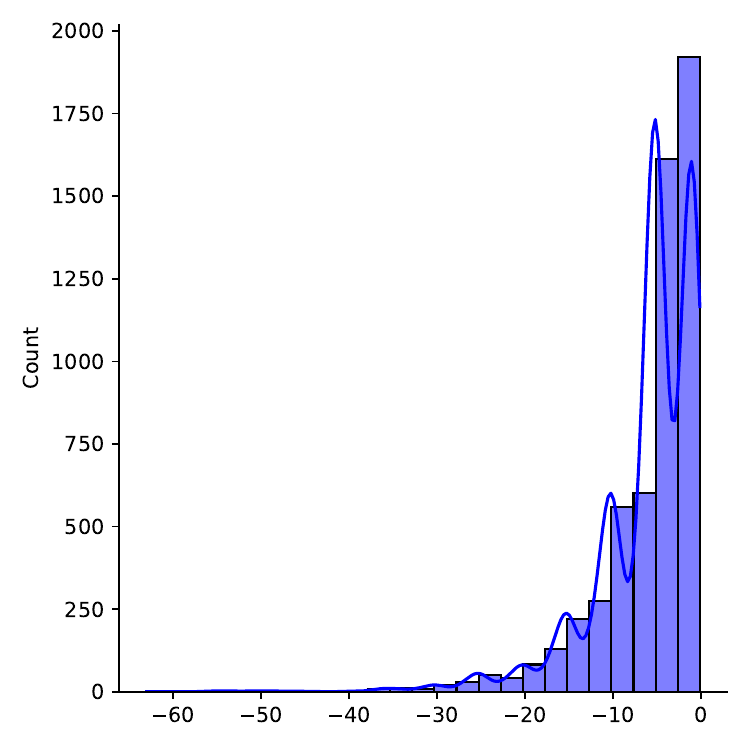}
            \caption[]%
            {en-es}    
            \label{sdd2}
        \end{subfigure}
        \vskip\baselineskip
        \begin{subfigure}[b]{0.42\textwidth}   
            \centering 
            \includegraphics[width=\textwidth]{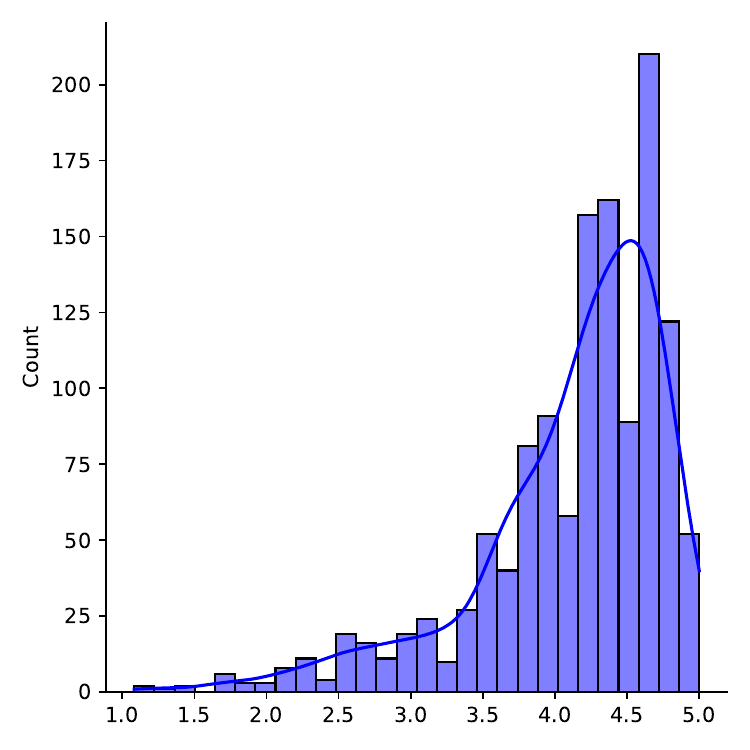}
            \caption[]%
            {en-zh}    
            \label{sdd3}
        \end{subfigure}
\caption{\cll{MQM and summarization score} distributions of our dev set. \cll{These scores are constructed with heuristics from human annotations. The annotation process is described in \citet{freitag-etal-2021-experts, freitag-etal-2022-results} and \citet{fabbri-etal-2021-summeval}.}  The blue line indicates the kernel density estimation describing the distribution.}
\label{score_distribution_dev}
\end{figure*}

Figure \ref{score_distribution_dev} we shows the distribution of scores on our dev-set.

\end{document}